\documentclass[sigconf]{acmart}  

\usepackage{booktabs} 

\newcommand{\Tref}[1]{Table~\ref{#1}}
\newcommand{\Eref}[1]{Eq.~(\ref{#1})}
\newcommand{\Fref}[1]{Fig.~\ref{#1}}
\newcommand{\Aref}[1]{Algorithm~\ref{#1}}
\newcommand{\Sref}[1]{Sec.~\ref{#1}}
\DeclareMathOperator*{\argmin}{arg\,min}

\graphicspath{{./img/}}
\usepackage{subfig}
\usepackage[ruled, vlined, linesnumbered]{algorithm2e} 
\usepackage{multirow} 
\usepackage{wrapfig} 

\usepackage{booktabs} 

\usepackage{enumitem}  

\newcommand{\bvec}[1]{\mbox{\boldmath $#1$}}
\newcommand{\bvecsmall}[1]{\mbox{\scriptsize \boldmath $#1$}}

\definecolor{DarkOliveGreen}{rgb}{0.33, 0.42, 0.18}
\newcommand{\codecommentcolor}{DarkOliveGreen} 

\newcommand{\codecomment}[1]{\textcolor{\codecommentcolor}{\texttt{#1}}}

\usepackage{balance}

\setcopyright{rightsretained}

\begin{document}
\title{PQk-means: Billion-scale Clustering for Product-quantized~Codes}

\author{Yusuke Matsui}
\authornote{Joint first authors.}
\affiliation{\institution{National Institute of Informatics, Japan}}
\email{matsui@nii.ac.jp}

\author{Keisuke Ogaki$^*$}
\affiliation{\institution{DWANGO Co., Ltd., Japan}}
\email{keisuke_ogaki@dwango.co.jp}

\author{Toshihiko Yamasaki}
\affiliation{\institution{The University of Tokyo, Japan}}
\email{yamasaki@hal.t.u-tokyo.ac.jp}

\author{Kiyoharu Aizawa}
\affiliation{\institution{The University of Tokyo, Japan}}
\email{aizawa@hal.t.u-tokyo.ac.jp}

\renewcommand{\shortauthors}{Y. Matsui et al.}

\begin{abstract}
Data clustering is a fundamental operation in data analysis.
For handling large-scale data, 
the standard k-means clustering method is not only slow, but also memory-inefficient.	
We propose an efficient clustering method for billion-scale feature vectors, called {\it PQk-means}.
By first compressing input vectors into short product-quantized (PQ) codes,
PQk-means achieves fast and memory-efficient clustering, even for
high-dimensional vectors.
Similar to k-means, PQk-means repeats the assignment and update steps,
both of which can be performed in the PQ-code domain.
Experimental results show that even short-length (32 bit) PQ-codes can produce competitive results compared with k-means.
This result is of practical importance for clustering in memory-restricted environments.
Using the proposed PQk-means scheme, the clustering of one billion 128D SIFT features with $K = 10^5$ is achieved
within 14 hours, using just 32 GB of memory consumption on a single computer.

\end{abstract}

%
%




\copyrightyear{2017}
\acmYear{2017}
\setcopyright{acmcopyright}
\acmConference{MM '17}{October 23--27, 2017}{Mountain View, CA, USA}\acmPrice{15.00}\acmDOI{10.1145/3123266.3123430}
\acmISBN{978-1-4503-4906-2/17/10}

\maketitle

\section{Introduction}
\label{sec:intro}
Many recent advances in computer vision are attributed to supervised learning with several annotated data sources.
However, manual annotation is a time-consuming and laborious task.
Clustering (unsupervised learning) is a promising method for taking better advantage of unlabeled data ~\cite{cvpr_yang2016}.
Specifically, we focus on million- or billion-scale clustering for data with hundreds or thousands of dimensions,
e.g., clustering on 100 million images with 4096D AlexNet features (YFCC100M~\cite{cacm_thomee2016}).

The problems of large-scale clustering include large memory consumption
and prohibitive runtime costs.
Owing to these two issues,
the standard k-means clustering method~\cite{tit_lloyd1982} can barely handle large-scale data.
Distributed batch clustering~\cite{www_sculley2010, nips_newling2016}
is a possible solution for achieving large-scale clustering within a reasonable timescale.
However, this requires vast computational resources.
For example, clustering 100 million features within several hours
requires 300 machines~\cite{iccv_avrithis2015} in a Spark framework\footnote{\url{http://spark.apache.org/}}.

In this paper, we propose \textit{PQk-means}, which is a \textbf{billion-scale} clustering method, and can be performed on a \textbf{single computer}
with only a \textbf{reasonable memory consumption} (less than 32 GB of RAM) \textbf{within a single day}.
The key idea is to first compress input vectors into memory-efficient short codes by product quantization~\cite{tpami_jegou2011},
and to then cluster the resultant product-quantized (PQ) codes (rather than the original vectors) in the compressed domain.
As with k-means, PQk-means also repeats the following two steps until convergence is achieved:
(1) Find the nearest center from each code,
and (2) update each center using a proposed \textit{sparse voting} scheme.

\begin{figure}
	\begin{center}
		\subfloat[k-means]{\includegraphics[width=0.5\linewidth]{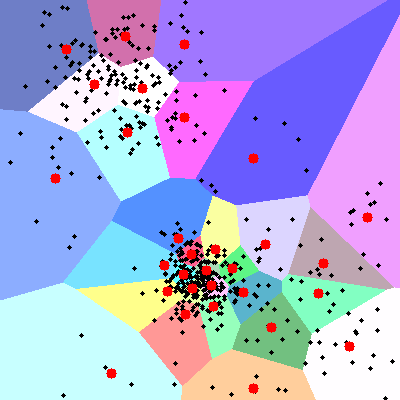}
			\label{fig:teaser_kmeans}}
		\subfloat[PQk-means]{\includegraphics[width=0.5\linewidth]{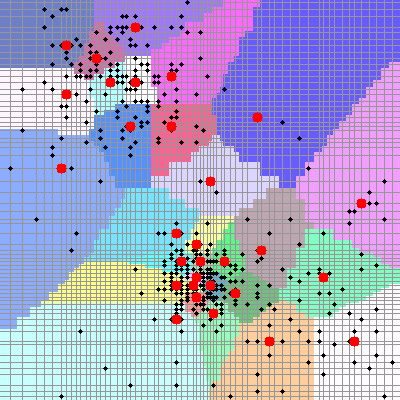}
			\label{fig:teaser_pqkmeans}}
	\end{center}
		\vspace{-2mm}
	\caption{
				A 2D example using both k-means and PQk-means, with $K=30$.
				(a) K-means applied to 500 2D vectors (black dots), and the resulting 30 centers denoted as red circles.
				(b) The same 500 vectors encoded as 500 PQ codes (black dots), and the resulting 30 centers using PQk-means.
				Because both dimensions ($x$ and $y$) are quantized independently, the PQ codes are placed on nonuniformly quantized lattices.}
		
	\label{fig:teaser}
\end{figure}

\Fref{fig:teaser} illustrates a 2D example, comparing k-means with PQk-means.
The result of the PQk-means procedure is similar to that of k-means, although PQk-means is 5.3 times more memory efficient\footnote{For representing each dimension, a 32 bit \texttt{float} is used in k-means, and a 6 bit integer is used in PQk-means ($2^6=64$ codewords are used).}.

The technical challenge lies in the direct computation of the center of a new cluster.
Because PQ codes consist exclusively of sets of identifiers (integers),
averaging operations on such codes cannot be explicitly defined.
A na\"{i}ve brute-force updating method is slow, because all possible candidates must be evaluated.
To solve this problem, we develop an alternative fast method, called \textit{sparse voting}.
We consider a frequency histogram of the assigned PQ codes in each cluster.
Owing to the nature of clustering, this histogram is usually sparse.
By focusing only on non-zero elements in the histogram, we can omit most calculations.
Sparse voting is a simple procedure. However, it significantly accelerates the computation, and achieves exactly the same results as the na\"{i}ve method.

We analyzed the runtime and memory consumption of PQk-means.
Moreover, we intensively compared PQk-means with
existing methods, such as k-means~\cite{tit_lloyd1982}, Bk-means~\cite{cvpr_gong2015},
Ak-means~\cite{cvpr_philbin2007}, and IQ-means~\cite{iccv_avrithis2015},
using the SIFT1M and ILSVRC\_1000C datasets. 
Billion-scale evaluation was also investigated, using the YFCC100M, SIFT1B, and Deep1B datasets.

The contributions of this paper are as follows:
\begin{itemize}[leftmargin=20pt]
	\item We develop PQk-means, a billion-scale memory-efficient clustering algorithm.
	The clustering of one billion 128D SIFT vectors with $K=10^5$ was achieved in 14 hours, using just 32 GB of RAM.
	Note that standard k-means clustering requires 512 GB of RAM just to represent the data.
	\item PQk-means is conceptually simple, and straightforward to implement.
	\item Unlike existing large-scale clustering methods
	such as Bk-means~\cite{cvpr_gong2015} or IQ-means~\cite{iccv_avrithis2015},
	the original vectors can be approximately reconstructed following the clustering.
	This is a useful property if the original vectors are required after clustering.
	\item Experimental results show that clustering with short-length PQ codes (e.g., 32 bit)
	is still effective
	(see \Fref{fig:cluster_example} for visual examples).
	This is a practically important result for memory-efficient clustering.
	
\end{itemize}

\section{Related work}
Data clustering is a fundamental operation in data analysis~\cite{pr_jain2010,assp_gray1984}.
Since the original k-means clustering method was proposed~\cite{tit_lloyd1982},
several theoretical improvements have been presented.
In particular, the provision of good seeds~\cite{soda_arthur2007, aaai_bachem2016, nips_bachem2016} and
bounding-based acceleration~\cite{icml_elkan2003, icml_newling2016} are still being
intensively studied.
Because these algorithms are based on k-means, 
they always produce the same final clustering result with the same initial seed~\cite{icml_newling2016}.

Considering real-world applications, faster algorithms are in high demand,
even though the result of clustering in such cases is only an approximation of that of k-means.
Such approximated k-means methods include approximated search~\cite{cvpr_philbin2007},
hierarchical search~\cite{cvpr_nister2006}, approximated bounds~\cite{cvpr_wang2012b},
and batch-based methods~\cite{www_sculley2010, nips_newling2016}.
If the size of the input data is large, subset-based methods~\cite{wsdm_broder2014, iccv_avrithis2015}
can achieve the fastest performance.
These methods only treat a subset of the input vectors (i.e., vectors close to each center),
making the computation efficient.
The current state-of-the-art 
for subset-based methods is IQ-means~\cite{iccv_avrithis2015}.
While subset-based methods are fast, their accuracy is not always competitive compared with other methods, because only subsets of vectors are used.

For handling large-scale data (e.g., $10^8$ 4096D AlexNet features, which require 1.6 TB in total using \texttt{float}), memory consumption also constitute an important issue.
Binary k-means (Bk-means)~\cite{cvpr_gong2015} converts input vectors into binary codes~\cite{tpami_gong2013, mm_dai2016},
so that all of the binary codes can be stored in memory.
The Hamming distance between two binary codes approximates the Euclidean distance between their original vectors.
Because the Hamming distance can be computed efficiently by either a linear scan or hash table~\cite{tpami_norouzi2014, cvpr_ong2016},
Bk-means achieves fast clustering with efficient memory utilization.
The drawbacks of Bk-means are two-fold.
First, binary conversion is less accurate than quantization-based compression~\cite{cvpr_babenko2014,cvpr_babenko2015,tpami_ge2014,cvpr_norouzi2013,tkde_wang2015,icml_zhang2014,cvpr_zhang2015, eccv_douze2016},
as has been discussed in the nearest neighbor search community~\cite{tpami_wang2017}.
Second, we cannot reconstruct the original vectors from the resultant binary codes.
Our PQk-means method addresses these two concerns. 
Experimental results show that PQk-means always achieves a better accuracy than Bk-means with the same code length. 
A comparison of standard k-means, Bk-means, and PQk-means is summarized in~\Tref{tbl:method_cmp}.

\begin{table*}
	\begin{center}
		\caption{
			Comparison of k-means, Bk-means, and the proposed PQk-means clustering methods.
			In k-means clustering, the vector is a $D$-dimensional real-valued vector. In Bk-means clustering, the vector is a $B$-bit binary string.
			In PQk-means clustering, the vector is a tuple consisting of $M$ indices, whose range is from $1$ to $L$.}
		\label{tbl:method_cmp}
		\begin{tabular}{@{}llll@{}} \toprule
			Method & Representation & Step1: Assignment & Step2: Updating \\ \midrule
			k-means~\cite{tit_lloyd1982} & $\bvec{x} \in \mathbb{R}^D $ & Nearest neighbor search & Averaging \\
			Bk-means~\cite{cvpr_gong2015} & $\bvec{x}\mapsto\bvec{x}_b \in \{0, 1\}^B$ & Hash table for binary codes~\cite{tpami_norouzi2014} & Bit operation~\cite{cvpr_gong2015} \\
			PQk-means (proposed) & $\bvec{x}\mapsto \bvec{\bar{x}} \in \{1, \dots, L \}^M $ & Hash table for PQ codes~\cite{iccv_matsui2015} (\Sref{sec:pqkmeans_assignment}) & Sparse voting (\Sref{sec:pqkmeans_update}) \\ \bottomrule
		\end{tabular}
		
		\vspace{-2mm}
	\end{center}
\end{table*}

\section{Background}
In this section, we briefly review k-means~\cite{tit_lloyd1982}
for clustering, and product quantization~\cite{tpami_jegou2011} for encoding.

\subsection{K-means clustering}
The k-means algorithm finds $K$ cluster centers such that the sum of the distances between each vector and its closest center is minimized. 
Specifically, given $N$ $D$-dimensional vectors $\mathcal{X}=\{\bvec{x}_n \in \mathbb{R}^D \}_{n=1}^N$,
one must find $K$ centers $\{ \bvec{\mu}_k \in \mathbb{R}^D \}_{k=1}^K$ that minimize the following cost function~\cite{tit_lloyd1982, nips_newling2016}:
\begin{equation}
	E \left (\bvec{\mu}_1, \dots, \bvec{\mu}_K  \right )= \frac{1}{N} \sum_{n=1}^N  d\left ( \bvec{x}_n, \bvec{\mu}_{a(n)} \right ),
	\label{eq:kmeans_obj}
\end{equation}
where $d(\bvec{x}, \bvec{y}) = \Vert \bvec{x} - \bvec{y} \Vert_2 $.
Note that $a(n)$ is an assignment function, defined by 
$a(n)=\argmin_{k \in \{1, \dots, K\}}d(\bvec{x}_n, \bvec{\mu}_k)^2$.

The cost function converges to a local minimum by repeating the following two steps.
In the \textbf{assignment step}, each vector is assigned to the nearest center.
This is achieved by computing $a(n)$ for each $n \in \{1, \dots, N\}.$
In the \textbf{update step}, each center is updated 
by averaging over the assigned vectors,
$\bvec{\mu}_k \gets \frac{1}{|\mathcal{X}_k|} \sum_{\bvecsmall{x} \in \mathcal{X}_k} \bvec{x}$,
where $\mathcal{X}_k = \{ \bvec{x}_n \in \mathcal{X} | a(n)=k \}$.

\subsection{Product quantization for encoding}
\label{sec:pq}
The product-quantization algorithm encodes input vectors into short codes~\cite{tpami_jegou2011}.
A $D$-dimensional input vector $\bvec{x}\in \mathbb{R}^D$ is split into $M$ disjointed subvectors.
For each $D/M$-dimensional subvector, the closest codeword from the pre-trained $L$ codewords is determined, and its index (an integer in $\{1, 2, \dots, L\}$) is recorded.
Finally, $\bvec{x}$ is encoded as $\bvec{\bar{x}}$, which is a tuple of $M$ integers defined as follows:
\begin{equation}
	\bvec{x} \mapsto \bvec{\bar{x}} = [\bar{x}^1, \dots, \bar{x}^M ]^\top \in \{1, \dots, L\}^M,
\end{equation}
where the $m$th subvector in $\bvec{x}$ is quantized into $\bar{x}^m$.
We refer to $\bvec{\bar{x}}$ as a PQ code for $\bvec{x}$.
Note that $\bvec{\bar{x}}$ is represented by $M\log_2 L$ bits.
We set $L$ to 256, in order to represent each code using $M$ bytes. This is a typical setting in many studies.

Note that for each subspace,
$L$ codewords are trained beforehand.
Therefore, we can compute a distance matrix among codewords for each subspace,
$A^m \in \mathbb{R}^{L \times L}$ for each $m \in \{ 1, \dots, M\}$, where $A^m_{i, j}$ denotes the squared distance
between the $i$th and $j$th codewords for the $m$th subspace.

Suppose we have two vectors, $\bvec{x}_1$ and $\bvec{x}_2$,
and that their PQ codes are $\bvec{\bar{x}}_1$ and $\bvec{\bar{x}}_2$, respectively.
Then, the Euclidean distance between $\bvec{x}_1$ and $\bvec{x}_2$ 
is efficiently approximated with the two codes $\bvec{\bar{x}}_1$ and $\bvec{\bar{x}}_2$;
$d(\bvec{x}_1, \bvec{x}_2) \sim d_{SD}(\bvec{\bar{x}}_1, \bvec{\bar{x}}_2)$.
This is known as the symmetric distance~(SD)~\cite{tpami_jegou2011}:
\begin{equation}
	d_{SD}(\bvec{\bar{x}}_1, \bvec{\bar{x}}_2)^2 = \sum_{m=1}^M d_{SD}^m(\bar{x}_1^m, \bar{x}_2^m)^2 = \sum_{m=1}^M A^m_{\bar{x}_1^m, \bar{x}_2^m}.
	\label{eq:sdc}
\end{equation}
The SD approximates the distance between the original vectors by the distance between codewords, denoted by PQ codes.
Here, $d_{SD}^m(i, j)$ computes the distance between the $i$th and $j$th codewords in the $m$th space,
and can be computed simply by looking up $A^m_{i, j}$.
Therefore, the squared SD can be efficiently computed using look-up tables with a summation of the results. This computation requires a cost of $O(M)$.

A useful property of product quantization is its reconstructability. Given a PQ code $\bvec{\bar{x}}$, an original vector $\bvec{x} \in \mathbb{R}^D$
can be approximately reconstructed by fetching the codewords $\bvec{\bar{x}} \mapsto \bvec{\hat{x}} \in \mathbb{R}^D$, where $\bvec{\hat{x}}$ is an approximation of $\bvec{x}$.

\section{PQk-means clustering}
\label{sec:pqkmeans}
In this section, we present our proposed PQk-means clustering method. 
We assume that $D$-dimensional input vectors $\mathcal{X} = \{ \bvec{x}_n \}_{n=1}^N$
are encoded beforehand using product quantization, as $\mathcal{\bar{X}} = \{ \bvec{\bar{x}}_n \}_{n=1}^N$.
Our objective is to determine $K$ cluster centers that minimize the cost function:
\begin{equation}
	E(\bvec{\bar{\mu}}_1, \dots, \bvec{\bar{\mu}}_K)=\frac{1}{N} \sum_{n=1}^N d_{SD} \left ( \bvec{\bar{x}}_n, \bvec{\bar{\mu}}_{a(n)}   \right ),
	\label{eq:pqkmeans_obj}
\end{equation}
where $\bvec{\bar{x}}_n = [ \bar{x}^1_n, \dots, \bar{x}^M_n ]^\top \in \{1, \dots, L \}^M$.
Note that each cluster center $\bvec{\bar{\mu}}_k = [\bar{\mu}^1_k, \dots, \bar{\mu}^M_k]^\top$ $\in \{ 1, \dots, L \}^M$ is also a PQ code.
Here, \Eref{eq:pqkmeans_obj} differs from \Eref{eq:kmeans_obj} in two aspects. First, both input vectors and centers are PQ codes.
Second, the symmetric distance $d_{SD}$ is used to measure the distance between two PQ codes.

Similar to the standard k-means clustering method,
PQk-means repeats the assignment and update steps until convergence is achieved.

\subsection{Assignment step}
\label{sec:pqkmeans_assignment}
In the assignment step, the nearest center in terms of the SD is determined for each $\bvec{\bar{x}}_n$:
\begin{equation}
	a(n)=\argmin_{k \in \{1, \dots, K\}}d_{SD}(\bvec{\bar{x}}_n, \bvec{\bar{\mu}}_k)^2.
	\label{eq:pq_assign}
\end{equation}

There are two methods to compute \Eref{eq:pq_assign}: the PQ linear scan~\cite{tpami_jegou2011} or PQTable~\cite{iccv_matsui2015}.
For each $\bvec{\bar{x}}_n$, the PQ linear scan simply retrieves the closest of the $K$ centers
$\{\bvec{\bar{\mu}}_{k}\}_{k=1}^K$ linearly using \Eref{eq:sdc}.
This computation requires a cost of $O(KM)$ for each $\bvec{\bar{x}}_n$.
This is sufficiently fast for a small $K$ value, but is not efficient if $K$ is large.
The PQTable is a hash-table-based acceleration data structure, which is tailored for the efficient computation of $d_{SD}$.
When the number of items ($K$) is small, the computational cost of managing and hashing the PQTable is larger than for the PQ linear scan.
However, for large $K$ values, the PQTable is between $10^2$ and $10^5$ times faster than the PQ linear scan~\cite{iccv_matsui2015}.

Given the input PQ codes $\mathcal{\bar{X}}$,
it is not easy to decide which method to use, because the computational cost depends on $K$, $M$, and the distribution of vectors of the target dataset.
We adopt a simple but effective approach.
Given the PQ codes, we first evaluate both methods several times, and then select the faster of the two.
We found that this simple selection method is also useful for Bk-means, and therefore we incorporated this technique into the Bk-means method for the evaluation.

\subsection{Update step}
\label{sec:pqkmeans_update}
Once each input PQ code is assigned to its nearest cluster center,
we update each cluster center such that the sum of the errors within the cluster is minimized.
For typical real-valued vectors,
this can be achieved by computing the mean vector among all of the vectors in each cluster.
However, no method is known for computing a ``mean PQ code'' from a set of PQ codes.
Here, we define the mean PQ code as that which minimizes the sum of the symmetric distances to each PQ code within a cluster.

We can propose a na\"{i}ve straightforward method.
The na\"{i}ve method is a brute-force approach, which is therefore slow.
The experimental results show that this na\"{i}ve method is sometimes even slower than
the assignment step, as we will discuss in \Sref{sec:exp_runtime}.
Consequently, we develop an alternative method, called \textit{sparse voting}.
By reorganizing the items in the cluster, sparse voting achieves the same result as the na\"{i}ve method, but more efficiently.
This simple modification accelerates the computation significantly (10$\times$ to 50$\times$).

\textbf{Na\"{i}ve method:} Let us focus on the $k$th cluster.
For simplicity, we refer to the PQ codes assigned to the cluster as $\{\bvec{\bar{x}}_n \}_{n=1}^{N_k}$,
where $N_k \sim N / K$.
The purpose here is to
compute a new center $\bvec{\bar{\mu}}_k=[\bar{\mu}_k^1, \dots, \bar{\mu}_k^M]^\top \in \{1, \dots, L \}^M$.
Because each subspace is independent,
we consider the $m$th subspace.
Therefore, the problem is defined as follows:
Given $N_k$ integers $\{ \bar{x}_n^m \}_{n=1}^{N_k}$, where each $\bar{x}_n^m \in \{ 1, \dots, L\} $,
we calculate the ``mean'' code
$\bar{\mu}_k^m \in \{ 1, \dots, L\} $.

The straightforward brute-force approach tests all possible candidates.
Subsequently, the best candidate that minimizes the sum of the errors within the cluster is determined as follows:
\begin{equation}
	\bar{\mu}_k^m \gets \argmin_{l \in \{ 1, \dots, L \}} \sum_{n=1}^{N_k} d_{SD}^m(\bar{x}_n^m, l)^2.
	\label{eq:naiive}
\end{equation}
Using \Eref{eq:sdc}, we find that $d_{SD}^m(\bar{x}_n^m, l)^2 = A^m_{\bar{x}_n^m, l}$.
Therefore, this can be computed by simply looking up the table.
This na\"{i}ve computation requires a cost of $O(LN_k)$.

\textbf{Sparse voting:}
Next, we develop a fast alternative method, called sparse voting.
By creating a histogram, we can efficiently compute \Eref{eq:naiive}.
Given $\{ \bar{x}_n^m \}_{n=1}^{N_k}$, we scan these, and create an $L$-dimensional histogram of frequency:
\begin{equation}
	\bvec{h} = [h_1, \dots, h_{L}]^\top \in \mathbb{N}^{L},
	\label{eq:hist}
\end{equation}
where $h_l$ denotes the frequency with which an integer $l$ appears in $\{ \bar{x}_n^m \}_{n=1}^{N_k}$.
This scanning process requires a cost of $O(N_k)$.

Using $\bvec{h}$, \Eref{eq:naiive} can be rewritten as
\begin{equation}
	\bar{\mu}_k^m \gets \argmin_{l \in \{ 1, \dots, L \}} v_l,~~\textrm{where}~[v_1, \dots, v_{L}]^\top = A^m\bvec{h}.
	\label{eq:error_voting}
\end{equation}
It is easy to show that the right-hand side of \Eref{eq:naiive} is equivalent to that of \Eref{eq:error_voting} once they are expanded.
The computational cost of \Eref{eq:error_voting} is $O(L^2)$.

Furthermore, if $\bvec{h}$ is sparse, then the cost becomes $O(L \| \bvec{h} \|_0)$,
where $\| \bvec{h} \|_0 \in \{ 0, \dots, L\}$ denotes the number of nonzero elements in $\bvec{h}$.
Thus, the entire cost of sparse voting is $O(N_k + L \| \bvec{h} \|_0)$.
Although sparse voting is a simple trick, it accelerates the computation significantly.

\textbf{Analysis:}
With both the na\"{i}ve method and the sparse voting method, the final center
$\bvec{\bar{\mu}}_k=[\bar{\mu}_k^1, \dots, \bar{\mu}_k^M]^\top$ is created by computing $\bar{\mu}_k^m$ for all $m$.
Therefore, for each cluster, the computational costs of the na\"{i}ve method and
the sparse-voting method are $O(LMN_k)$ and $O(M(N_k + L \| \bvec{h} \|_0))$, respectively.
Finally, by summing $K$ clusters, we find that the total costs are $O(LMN)$ and $O(M(N + KL \| \bvec{h} \|_0))$, respectively.

If the constant factor is the same, then sparse voting is faster when
$N / K > \frac{L}{L-1} \| \bvec{h} \|_0 \sim \| \bvec{h} \|_0$.
Because the PQ codes in the same cluster tend to be similar (owing to the nature of clustering),
the histogram $\bvec{h}$ tends to be sparse, and this condition is satisfied in many cases, as we discuss in \Sref{sec:exp_runtime}.

\subsection{Pseudocode}
\Aref{alg:pqkmeans} presents the pseudocode for PQk-means.
The pipeline is extremely simple.
The \texttt{Init()} function initializes the centers, by simply randomly picking up $K$ codes from the input codes.
The \texttt{Check()} function decides the manner in which the nearest neighbors are found,
whether by a PQ linear scan or with a PQTable. This can be achieved by simply running both methods 10 times with randomly sampled vectors.
\texttt{BuildTable()} creates a PQTable. 
\texttt{FindNN()} and \texttt{UpdateCenter()} are explained in \Sref{sec:pqkmeans_assignment} and \Sref{sec:pqkmeans_update},
respectively. The results of the assignment function $a(n)$ are stored in an array.
Any condition can be adopted as a stop condition.

\begin{algorithm}[t]
	\KwIn{$\mathcal{\bar{X}}=\{ \bvec{\bar{x}}_n \}_{n=1}^N$,\ \ \ \ \ \ \ \ \ \ \ \codecomment{// PQ codes}
		\\\ \ \ \ \ \ \ \ \ \ \ \ $\mathcal{A}=\{ A^m \}_{m=1}^M$,\ \ \ \ \ \ \ \ \codecomment{// Distance matrices}
		\\\ \ \ \ \ \ \ \ \ \ \ \ $K.$\ \ \ \ \ \ \ \ \ \ \ \ \ \ \ \ \ \ \ \ \ \ \ \ \ \ \ \ \ \codecomment{// The number of clusters}}
	\KwOut{$\mathcal{\bar{M}}=\{ \bvec{\bar{\mu}}_k \}_{k=1}^K$.\ \ \ \ \ \ \codecomment{// PQ codes}}
	$\mathcal{\bar{M}} \gets$ Init($\mathcal{\bar{X}}$) \\
	$flag \gets$ Check($\mathcal{\bar{X}}, \mathcal{A}$) \\
	\Repeat{stop condition}{
		$a \gets \emptyset$ \ \ \ \ \ \ \ \ \codecomment{// Array}\\
		\If{$flag$}{		
			$table \gets$ BuildTable$(\mathcal{\bar{M}}, \mathcal{A})$ \ \ \ \codecomment{// PQTable}\\
		}
		\For{$n \gets 1~\mathrm{to}~N$}{
			\uIf{$flag$}{
				\codecomment{// \Sref{sec:pqkmeans_assignment} (PQTable)} \\
				$a[n] \gets$ FindNN$(\bvec{\bar{x}}_n, table)$ 
			}
			\Else{
				\codecomment{// \Sref{sec:pqkmeans_assignment} (PQ linear scan)} \\
				$a[n] \gets$ FindNN$(\bvec{\bar{x}}_n, \mathcal{\bar{M}}, \mathcal{A})$ 
			}
		}
		\For{$k \gets 1~\mathrm{to}~K$}{
			$\bvec{\bar{\mu}}_k \gets $UpdateCenter$(\mathcal{\bar{X}}, a, \mathcal{A})$ \ \ \ \codecomment{// \Sref{sec:pqkmeans_update}} 
		}
	}
	\caption{PQk-means clustering}
	\label{alg:pqkmeans}
\end{algorithm}

\section{Experimental results}
\label{sec:exp}
We evaluated PQk-means using various datasets.
All experiments were performed on a server with 3.0 GHz Intel Xeon CPUs (4 cores, 8 threads) and 128 GB of RAM
\footnote{We verified that the largest experiment (\Sref{sec:exp_large_scale}) was also run on a computer with only 32 GB of RAM.}.
For a fair comparison with existing methods, we employed a single-thread implementation for clustering
(\Sref{sec:exp_runtime}, \ref{sec:exp_memory}, \ref{sec:exp_comp}, and \ref{sec:exp_comp_several}).
For large-scale clustering (\Sref{sec:exp_large_scale}), we used a multithread implementation, in order to highlight the best performance.
All source codes are publicly available on \texttt{https://github.com/DwangoMediaVillage/pqkmeans}.

\subsection{Setup}

\textbf{Compared methods:}
For comparison, we implemented standard k-means clustering~\cite{tit_lloyd1982},
Bk-means~\cite{cvpr_gong2015} with iterative quantization (ITQ)~\cite{tpami_gong2013},
and Ak-means~\cite{cvpr_philbin2007} using FLANN~\cite{tpami_muja2014}.
Ak-means is an accelerated version of k-means, where the assignment step is accelerated using a KD tree.
In addition, we compared our method with IQ-means~\cite{iccv_avrithis2015}, which is the latest subset-based method.
Note that k-means, Ak-means, and IQ-means are \textbf{not} memory efficient for high-dimensional vectors.

\textbf{Datasets and features:}
We used four datasets: ILSVRC2012, BIGANN, YFCC, and Deep1B. 
The details of each dataset are summarized in \Tref{tbl:datasets},
where \#test denotes the number of input vectors on which the clustering algorithms were applied.
Likewise, \#train denotes the number of vectors used for training the codewords for product quantization and the rotation matrices for ITQ.

\begin{table}
	\begin{center}
		\caption{
			Dataset statistics.
		}
		\label{tbl:datasets}				
		\begin{tabular}{@{}lrrr@{}} \toprule 
			Dataset & $D$ & \#train & \#test \\ \midrule	
			ILSVRC\_100C & 4,096 & 100K &  129,395 \\ 
			ILSVRC\_1000C & 4,096 & 100K &  1,281,167 \\ 
			SIFT1M & 128 & 100K &  1,000,000 \\
			SIFT1B & 128 & 1M &  1,000,000,000 \\
			YFCC100M & 4,096 & 2M &  96,419,740 \\		
			Deep1B & 96 & 1M & 1,000,000,000 \\ \bottomrule
		\end{tabular}
	\end{center}
	\vspace{-2mm}		
\end{table}

The ILSVRC2012 dataset is a subset of ImageNet~\cite{ijcv_russakovsky2015}.
This dataset consists of 1000 object categories, each of which contains around 1000 images.
According to Gong et al.~\cite{cvpr_gong2015}, the full dataset was named ILSVRC\_1000C, and a small subset named ILSVRC\_100C
which was constructed by randomly picking 100 classes.
For each image, we extracted a 4096D AlexNet feature~\cite{nips_krizhevsky2012},
which was activated from the last hidden layer, using
the chainer framework~\cite{lsys_tokui2015} with a pretrained model.
We used 100K test images 
for training\footnote{Because ILSVRC2012 is used for image-recognition competitions,
	it contains more training images than test images.
	However, as our objective is clustering, we reversed the two groups,
	using the training images as test images and vice versa.}.

From BIGANN~\cite{icassp_jegou2011}, we used the two datasets SIFT1M and SIFT1B.
For the training of SIFT1B, we used the top one million vectors from the whole training set.

Yahoo's Flickr Creative Commons 100M (YFCC100M)
dataset \cite{cacm_thomee2016} contains around 100M images. 
An AlexNet feature vector was extracted from each image, as with ILSVRC.
Two million randomly chosen features were used for training.

The Deep1B dataset~\cite{cvpr_babenko2016} contains one billion test and 350M training features.
Each feature was extracted from the last fully connected layer of GoogLeNet~\cite{cvpr_szegedy2015} for one billion images. 
The features were compressed to 96 dimensions using PCA, and $l_2$ normalized.
For training, we used the top 1M vectors from the training set.

We used ILSVRC\_100C, ILSVRC\_1000C, and SIFT1M to compare the methods.
Note that each dataset has a distinct nature. The AlexNet features have a larger dimension and a sparse nature,
whereas the SIFT features are dense and structured.
The datasets YFCC100M, SIFT1B, and Deep1B were used for the large-scale evaluation.
Because the YFCC100M dataset includes the original images, 
the results of image clustering are evaluated visually (this will be illustrated later in \Fref{fig:cluster_example}).

\textbf{Encoding:}
For feature encoding, we employed PQ~\cite{tpami_jegou2011} for PQk-means, and
ITQ~\cite{tpami_gong2013} for Bk-means~\cite{cvpr_gong2015}.
The PQ codewords and ITQ rotation matrices were trained beforehand, using the training datasets.
Subsequently, all of the features were converted to $B$-bit PQ codes for PQ, and $B$-bit binary strings for ITQ, where $B=32$, $64$, and $128$. 
Note that $B=M\log_2 L=8M$ for the PQ codes.
Hereafter, we employ abbreviations to denote encodings with various bit lengths, e.g., 
``pqkmeans32'' refers to 32-bit PQ encoding.
Note that the bit length is a parameter specified by the user.
Larger bit lengths improve the accuracy, but require more memory.

For each vector, the encoding of ITQ requires a cost of $O(D^2)$, and that of PQ requires $O(DL)$.
The actual runtime of the encoding using ILSVRC\_1000C was 522 s for ITQ, and 109 s for PQ.
As discussed in \Sref{sec:intro}, our assumption is that users only store encoded codes. Therefore, encoding is the preprocessing step used in this study.
Note that IQ-means also requires a similar encoding process.

\textbf{Seed:}
For the initial seeds of the clustering, we randomly sampled $K$ vectors from the input dataset.
We fixed seeds for all methods using the same conditions (dataset, $B$, and $K$), to ensure a fair comparison.
Note that in our preliminary study, we observed that the selection of seeds did not significantly affect the results 
\footnote{For example, ten trials showed that the mean error is 242 and the standard deviation is 0.12 for SIFT1M with $K=10^3$ using 32-bit codes.}.

\subsection{Runtime analysis}
\label{sec:exp_runtime}
We evaluated the runtime of the proposed PQk-means clustering method.
\Tref{tbl:time_assign_update} presents a runtime comparison for each step in the assignment,
the update using the na\"{i}ve method, and the update using the proposed sparse-voting scheme.
The results confirm the following points.
First, \textbf{the proposed sparse voting method is always faster than the na\"{i}ve updating method, with a large margin}
(e.g., 54$\times$ faster in ILSVRC\_1000C with $K=10^3$).
Second, \textbf{the na\"{i}ve updating method is sometimes even slower than the assignment step}
(for ILSVRC\_100C with $K=10^2$, ILSVRC\_1000C with $K=10^3$, and SIFT1M with $K=10^2$).

These results indicate that the proposed sparse-voting scheme is highly efficient,
even though it achieves the same accuracy as the na\"{i}ve method. 
Consequently, we employed sparse voting during the subsequent evaluations in this study.

Theoretically, the runtime of PQk-means is
\begin{equation}
	\min(O(KMN), O(NT_{table}))+O(M(N+KL \| \bvec{h} \|_0)).
\end{equation}
The first term corresponds to the assignment step, whether using a PQ linear scan ($O(KMN)$) or a PQTable ($O(NT_{table})$).
Note that $T_{table}$ denotes the cost of a search for nearest neighbors using the PQTable,
which is heavily dependent on the data distribution.
The second term corresponds to the updating step using the sparse-voting scheme.
Because of the efficiency
of updating cluster centers using the sparse voting scheme,
as shown in \Tref{tbl:time_assign_update},
the dominant step is the assignment.

Note that the runtime of PQk-means does not depend on the dimension $D$ of
the original vectors, meaning that our PQk-means method performs efficiently for high-dimensional vectors.

\begin{table}
	\begin{center}
		\caption{Runtime comparison for each step using 32-bit codes for various conditions.
			For each condition, we run PQk-means twice (w/ Na\"{i}ve or w/ sparse voting), and report the runtimes.
			For each condition, the most/least time-consuming step is highlighted using a bold/underlined font.
			We also report a macro average value of $\| \bvec{h} \|_0$.
			All values constitute averages over 20 iterations.}
		\label{tbl:time_assign_update}
		\scalebox{0.92}{
			\begin{tabular}{@{}llllll@{}} \toprule
				& & & & \multicolumn{2}{c}{Update [ms]} \\ \cmidrule(l){5-6}
				Dataset & $K$ & $\Vert \bvec{h} \Vert_0$ & Assignment [ms] & Na\"{i}ve & Sparse\\ \midrule
				\multirow{2}{*}{ {\small ILSVRC\_100C} } & $10^2$ & 64.0 & 91.1 & \textbf{301} & \underline{5.05} \\			
				& $10^3$ & 18.7 & \textbf{795} & 183 & \underline{10.1} \\ \midrule
				\multirow{3}{*}{ {\small ILSVRC\_1000C} } & $10^2$ & 133 & $1.09 \times 10^3$ & $\mathbf{4.72 \times 10^3}$  & \underline{124} \\
				& $10^3$ & 52.7 & $\mathbf{8.51 \times 10^3}$ & $5.29 \times 10^3$ & \underline{98.2}  \\
				& $10^4$ & 14.0 & $\mathbf{3.86 \times 10^4}$ & $2.03 \times 10^3$  & \underline{218} \\ \midrule 
				\multirow{2}{*}{ {\small SIFT1M} } & $10^2$ & 145 & 811 & $\mathbf{4.26 \times 10^3}$  & \underline{105} \\
				& $10^3$ & 77.1 & $\mathbf{6.21 \times 10^3}$ & $2.43 \times 10^3$  & \underline{103} \\ \bottomrule
			\end{tabular}
		}
	\end{center}
		\vspace{-3mm}
\end{table}

\begin{figure*}
	\def\time_error_width{0.33\linewidth}
	\begin{center}
		\subfloat[ILSVRC\_100C with $K=10^2$]{\includegraphics[width=\time_error_width]{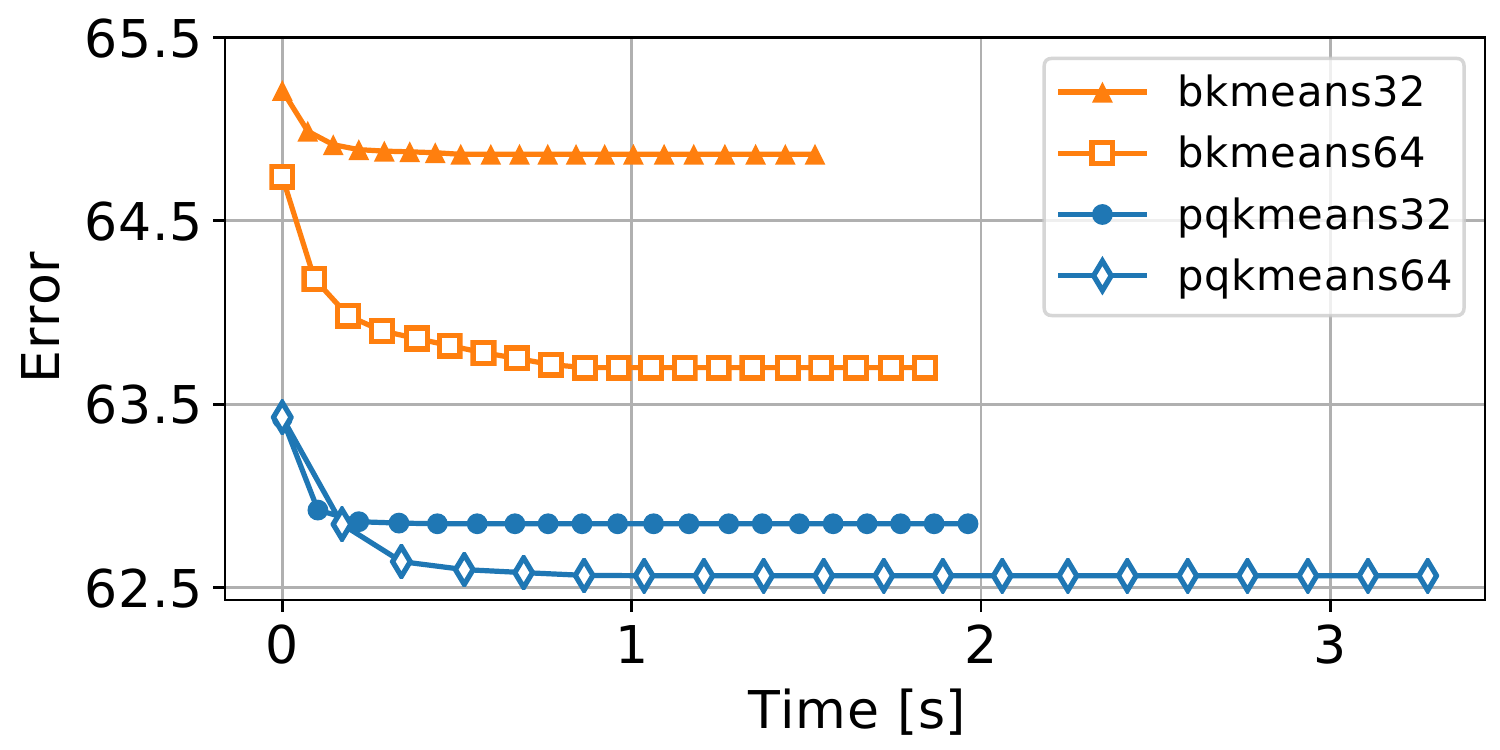}
			\label{fig:time_error_ilsvrc100c}}
		\subfloat[ILSVRC\_1000C with $K=10^3$]{\includegraphics[width=\time_error_width]{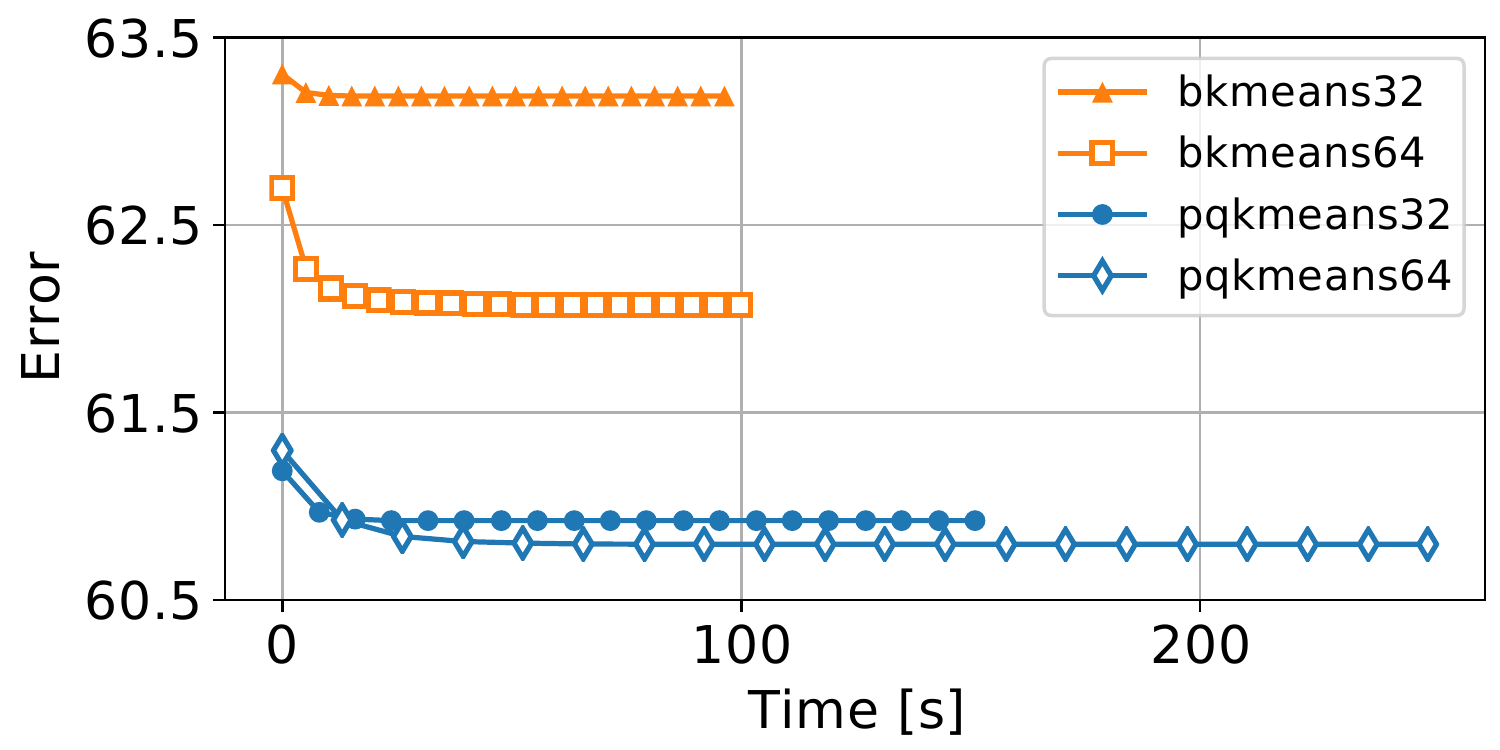}
			\label{fig:time_error_ilsvrc1000c}}
		\subfloat[SIFT1M with $K=50$]{\includegraphics[width=\time_error_width]{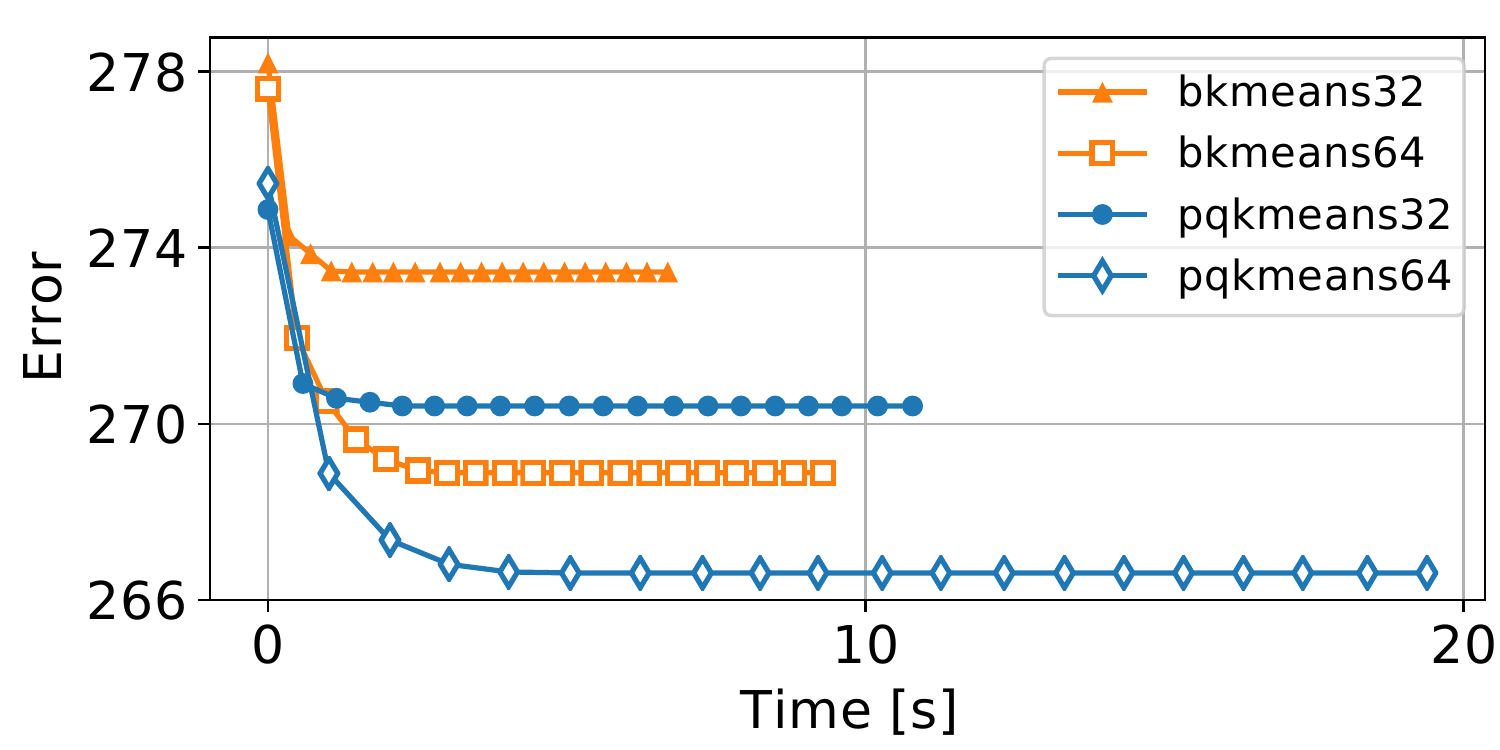}
			\label{fig:time_error_sift1m}}
	\end{center}
		\vspace{-2mm}
	\caption{Comparison of PQk-means with Bk-means in terms of errors and runtime. Errors are plotted for each iteration. All lines are plotted for 20 iterations. A relatively longer line indicates that more time was required.}
	\label{fig:time_error}
\end{figure*}

\subsection{Memory consumption}
\label{sec:exp_memory}
For $B$-bit codes, PQk-means requires $\frac{B}{8}(N+K)$ bytes for codes and centers, and $4L^2M$ bytes for distance matrices.
In addition, PQk-means requires an array of lists to specify an assignment (\textit{a} in \Aref{alg:pqkmeans}), which requires $4N$ bytes in total.
If the PQTable is used, this requires $4K \cdot 2^{Q(\log_2 (B/\log_2K))}$ bytes~\cite{iccv_matsui2015},
where $Q()$ is a rounding operation.
The sum mentioned above constitutes the theoretical runtime memory consumption.
Usually, $N$ is significantly larger than $K$. Therefore, \textbf{the main contributors to the memory are the input codes and an assignment array, $(B/8 + 4)N$}.

Compared to the standard k-means and the Ak-means methods, both of which require at least $4D(N+K)$ bytes for codes and centers,
the proposed PQk-means requires significantly less memory.
Because the memory consumption of the PQk-means does not depend on the dimension $D$ of the original vectors, 
PQk-means is particularly memory efficient when $D$ is large, such as for AlexNet features ($D=4096 $). 
For example, input vectors from ILSVRC\_1000C require $1,281,167 \times 4096 \times 4$ = 21 GB.
By contrast, PQ codes with 32 bits require only $1,281,167 \times 32 / 8$ = 5.12 MB.
This confirms the advantages of using short-code encoding schemes.
As shown in \Sref{sec:exp_comp}, even when features are encoded as very short codes,
the clustering performance declines slightly, with a substantial speed-up.
Notably, Bk-means offers a comparable advantage in terms of memory.

\subsection{Detailed comparison with Bk-means}
\label{sec:exp_comp}

We compared the proposed PQk-means method with Bk-means,
which is the closest comparable method (see \Tref{tbl:method_cmp}).
We examined the behavior of both methods at each iteration,
especially for relatively small $K$ values.
Because $K$ is small, the linear scan was used in the assignment step for both methods.
The results highlighted the general tendencies that PQk-means is more accurate, whereas Bk-means is faster.

Clustering errors were computed as follows.
Let us assume that either PQk-means or Bk-means is applied to the short codes to create $K$ clusters.
Following clustering, the corresponding \textbf{original} vectors $\{\bvec{x}_n \}_{n=1}^N$ are collected.
Subsequently, the error $E$ for the original vectors is computed using \Eref{eq:kmeans_obj}.
As $E$ measures the average errors in the original vectors (rather than codes),
we can compare the results of PQk-means using those of Bk-means.

Figure.~\ref{fig:time_error} presents the runtimes and errors during each iteration.
We obtained some interesting results.

\textbf{PQk-means vs. Bk-means}:
In the comparison of PQk-means with Bk-means for the same code length,
the former always achieved smaller errors.
This is because the employed product quantization is more accurate than ITQ, as reported in~\cite{cvpr_he2013}.
In terms of the runtime, Bk-means was always faster than PQk-means for the same code length.
This is because comparing bit strings is faster than comparing two PQ codes,
which also constitutes expected behavior~\cite{cvpr_he2013}.

\textbf{Code length}:
When considering different code lengths, there were smaller errors for longer bit lengths, as expected.
Interestingly, \textbf{the results for pqkmeans32 were more accurate than those of bkmeans64} in \Fref{fig:time_error_ilsvrc100c} and \Fref{fig:time_error_ilsvrc1000c}.
This could be explained by the higher expressiveness of PQ compared with that of ITQ.

\textbf{Convergence behavior}:
As can be observed, 20 iterations were sufficient to achieve convergence in all cases.
Note that if we stop the iteration when the error does not change from the previous iteration,
PQk-means can achieve similar computational cost as Bk-means for these datasets.

\subsection{Comparison with existing methods under several conditions}
\label{sec:exp_comp_several}

\begin{figure*}
	\def\k_n_error_time_width{0.25\linewidth}	
	\begin{center}
		\subfloat[\scriptsize{Runtime according to $K$.}]{\includegraphics[width=\k_n_error_time_width]{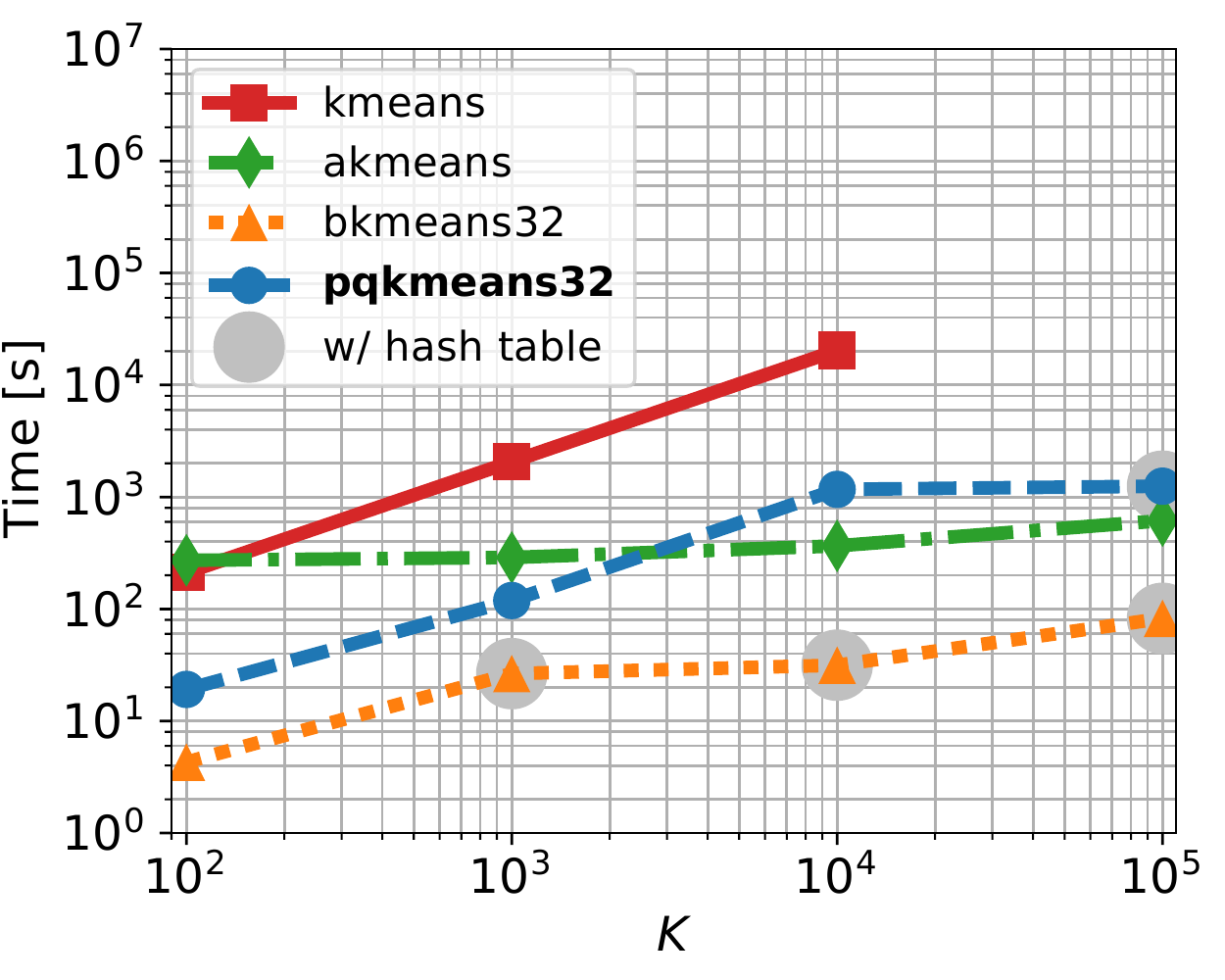}
			\label{fig:k_time}}
		\subfloat[\scriptsize{Runtime according to $N$ ($K=50$).}]{\includegraphics[width=\k_n_error_time_width]{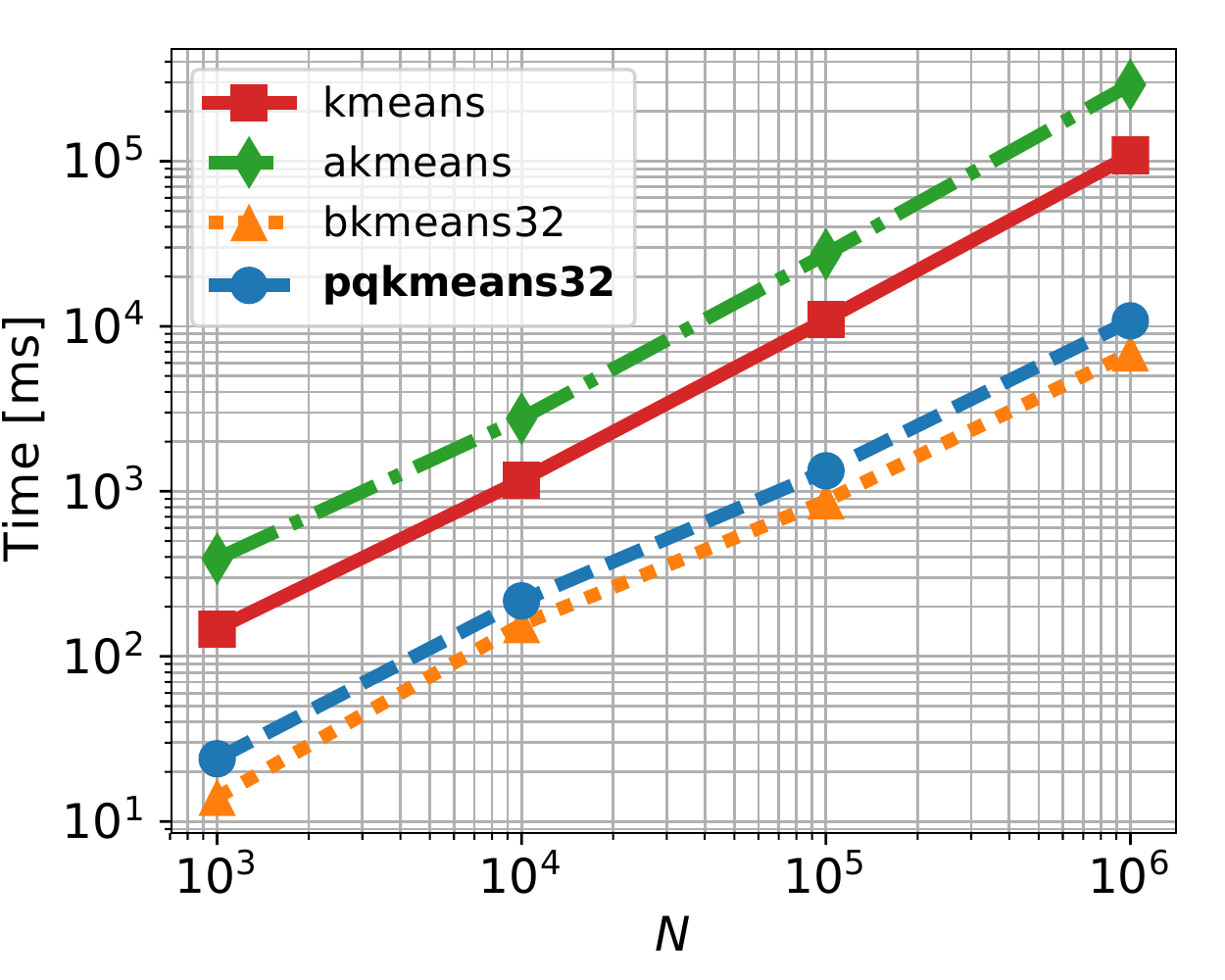}
			\label{fig:n_time}}
		\subfloat[\scriptsize{Error according to $N$ ($K=50$).}]{\includegraphics[width=\k_n_error_time_width]{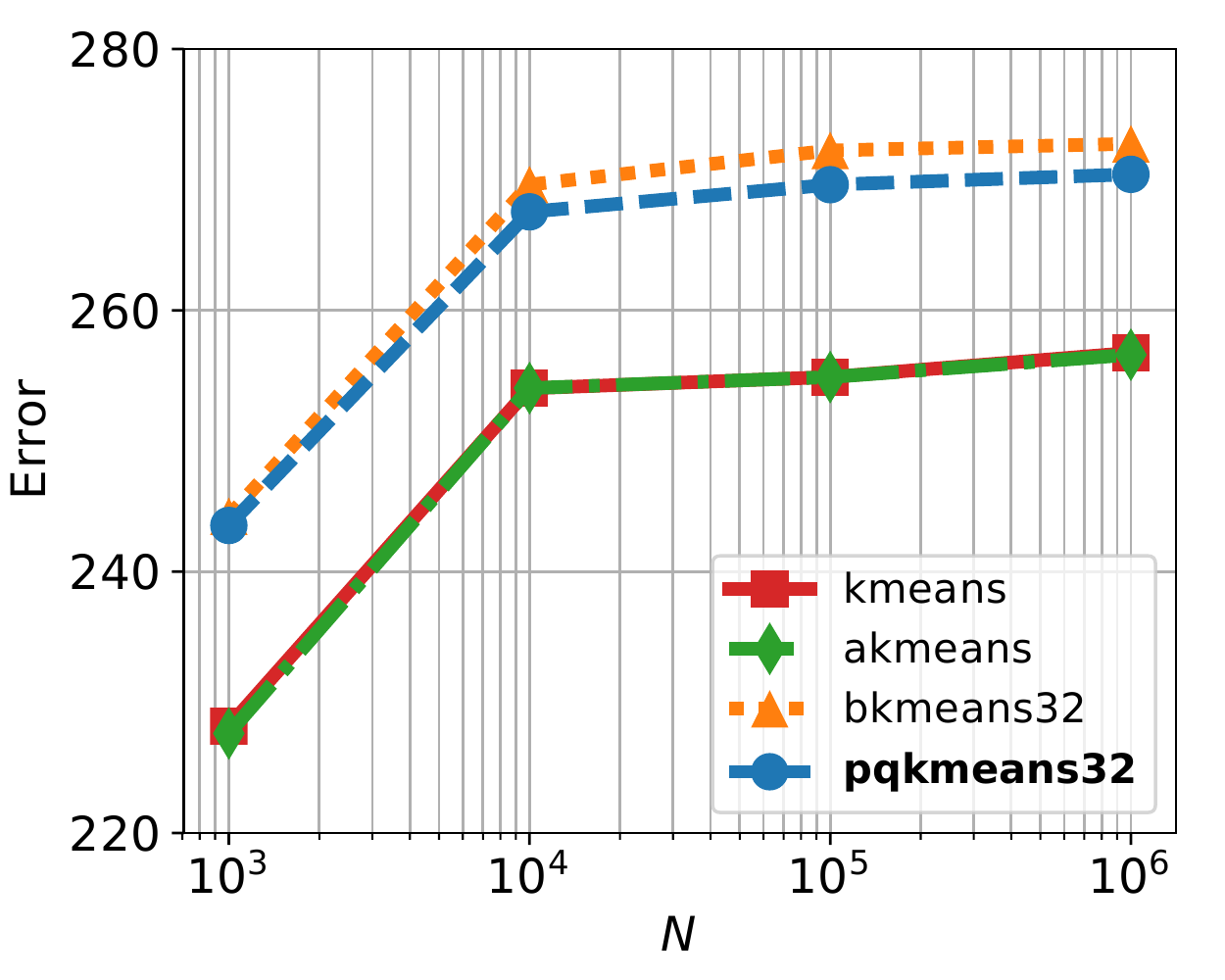}
			\label{fig:n_error}}
		\subfloat[\scriptsize{Error according to memory ($K=10^3$).}]{\includegraphics[width=\k_n_error_time_width]{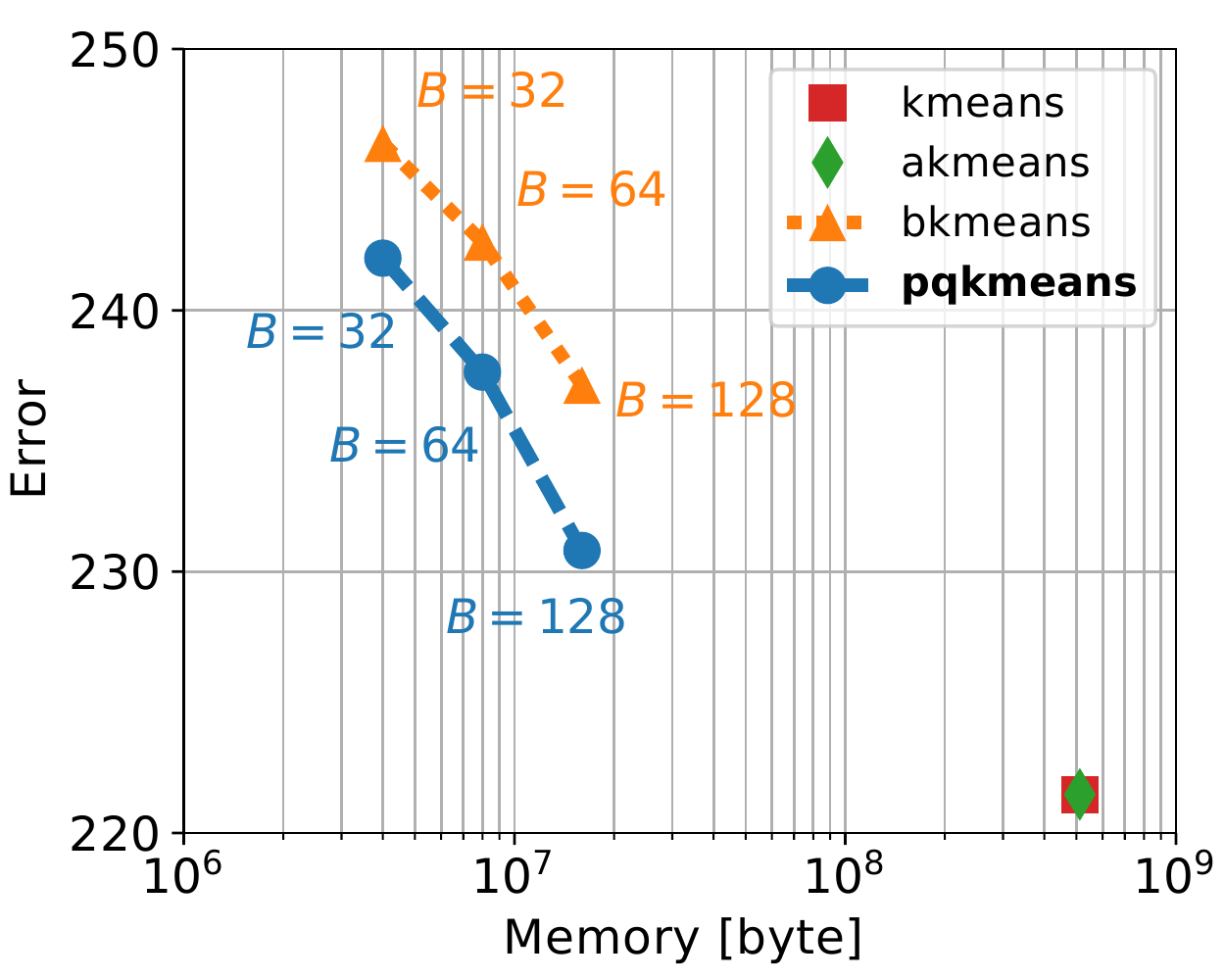}
			\label{fig:memory_error}}
		\qquad
	\end{center}
		\vspace{-2mm}
	\caption{Relation between the errors, runtime, and memory consumption of the input vectors, according to $N$ or $K$, for the SIFT1M dataset after 20 iterations.
		The gray dots in (a) indicate that hash tables (\cite{tpami_norouzi2014} for Bk-means and \cite{iccv_matsui2015} for PQk-means) were used in the assignment step.}
	\label{fig:k_n_error_time}
\end{figure*}

We compared the proposed PQk-means with Bk-means, k-means, and Ak-means under several conditions.
Our findings are summarized as follows:
\begin{itemize}[leftmargin=20pt]
	\item k-means was between 10$\times$ and 1,000$\times$ slower than PQk-means.
	\item Ak-means was accurate. However, it was slow for large $D$ and/or relatively small $K$ values.
	\item k-means and Ak-means required between 100$\times$ and 4,000$\times$ more memory than PQk-means.
	\item Short codes, such as 32-bit PQ codes, were effective in terms of the balance of accuracy, memory cost, and runtime.
\end{itemize}

\textbf{SIFT1M:}
Figure.~\ref{fig:k_n_error_time} illustrates the relationship between the runtime, errors, and memory consumption
according to $N$ or $K$ using SIFT1M.
As expected, k-means clustering resulted in the fewest errors in all cases (\Fref{fig:n_error}).
However, it was more than ten times slower in all cases compared with PQk-means and Bk-means (\Fref{fig:k_time}).

Figure.~\ref{fig:n_error} shows that Ak-means achieved low errors (almost the same as k-means).
However, owing to the overhead of the approximated search, Ak-means was slow for relatively
small $K$ (\Fref{fig:k_time}),
with Ak-means being slowest method for $K=50$ (\Fref{fig:n_time}).

Figure.~\ref{fig:memory_error} shows that Ak-means and k-means were not memory efficient,
even though these methods achieved lower errors.
Ak-means and k-means consumed 512 MB memory space for the vectors, 
whereas PQk-means and Ak-means required only $B/8$ MB.
\Fref{fig:memory_error} also shows that the error of PQk-means for 32-bit codes was lower than
that of Bk-means for 64-bit codes.

\textbf{ILSVRC\_1000C:}
A comparison using ILSVRC\_1000C with 32-bit codes is summarized in \Tref{tbl:exp_comp}.
To compare the results more intuitively, we introduce an additional evaluation criterion, the Rand index~\cite{jasa_rand1971}.
Given a pair of clustering results, the Rand index computes the similarity between them.
We compared the result of k-means against each method, where a higher Rand index indicates a higher similarity.

PQk-means was superior to Bk-means in terms 
the Rand index with the same code length.
For example, the Rand index of PQk-means (0.142) was higher than that of Bk-means (0.046) for $K=10^3$.

The errors of Ak-means were close to those of k-means. This was also confirmed by the high Rand index (e.g., 0.465 for $K=10^2$).
However, Ak-means required a huge amount of memory (21.0 GB, whereas 5.12 MB was required for PQk-means and Bk-means).
Moreover, because the runtime of Ak-means depends on the dimension of the vectors,
Ak-means was slower for high-dimensional features, such as AlexNet.
\Tref{tbl:exp_comp} shows that Ak-means was between 5$\times$ and 164$\times$ slower than PQk-means for all $K$.

Interestingly,
although the PQk-means with 32-bit codes was 20 times faster and required 4,000 times less memory than Ak-means,
the Rand index of PQk-means (0.142) is slightly lower than that of Ak-means (0.2) for $K=10^3$.
This implies that for the purpose of clustering,
the short-length code (e.g., 32-bit) can provide a strong balance between 
the accuracy, memory consumption, and runtime.
We believe that this is a practically important result for clustering in memory-restricted environments.

\begin{table}
	\begin{center}
		\caption{Comparison of methods using the ILSVRC\_1000C dataset with 32-bit codes after 20 iterations.}
		\label{tbl:exp_comp}				
		\begin{tabular}{@{}llllll@{}} \toprule
			Method & $K$ & Error & Rand index & Time [s] & Memory \\ \midrule
			\multirow{3}{*}{PQk-means} & $10^2$ & 65.09 & 0.230 & $18.2$ & \multirow{3}{*}{5.12 MB} \\
			& $10^3$ & 60.92 & 0.142 & $1.51 \times 10^2$  &  \\
			& $10^4$ & 59.03 & - & $7.22 \times 10^2$  &  \\ \midrule
			\multirow{3}{*}{Bk-means} & $10^2$ & 66.35 & 0.111 & $12.3$  & \multirow{3}{*}{5.12 MB} \\
			& $10^3$ & 63.19 & 0.046 & $1.00 \times 10^2$  &  \\
			& $10^4$ & 60.70 & - & $1.15 \times 10^2$  &  \\ \midrule
			\multirow{2}{*}{k-means} & $10^2$ & 64.25 & 1.0 & $9.12 \times 10^3$ & \multirow{2}{*}{21.0 GB} \\
			& $10^3$ & 58.95 & 1.0 & $1.06 \times 10^5$ &  \\ \midrule
			\multirow{3}{*}{Ak-means} & $10^2$ & 64.29 & 0.465 & $ 3.00 \times 10^3$  & \multirow{3}{*}{21.0 GB} \\
			& $10^3$ & 59.76 & 0.200 & $ 2.96 \times 10^3$  &  \\ 
			& $10^4$ & 56.78 & - & $ 3.65 \times 10^3$  &  \\ \bottomrule 
		\end{tabular}
	\end{center}
	\vspace{-2mm}
\end{table}

\subsection{Large-scale clustering evaluation}
\label{sec:exp_large_scale}
In this section, we present the results of a large-scale evaluation using
three billion-scale datasets, namely the YFCC100M, 
SIFT1B, 
and Deep1B datasets.
For the three datasets, we ran PQk-means with 32-bit codes and
various values of $K$ using a parallel implementation on a single machine.
In addition, we ran Bk-means with a parallel implementation for YFCC100M.
To highlight the best performance, we stopped the iteration when the error converged. 
The number of iterations required for convergence was five for all datasets.
Because these datasets are extremely large (1.58 TB, 512 GB, and 384 GB, for YFCC100M, SIFT1B, and Deep1B, respectively),
an ordinary computer cannot store all of the original data in its memory simultaneously.

\begin{figure}
	\def\large_exp_yfcc100m_width{0.5\linewidth} 
	\begin{center}
		\subfloat[Error according to $K$]{\includegraphics[width=\large_exp_yfcc100m_width]{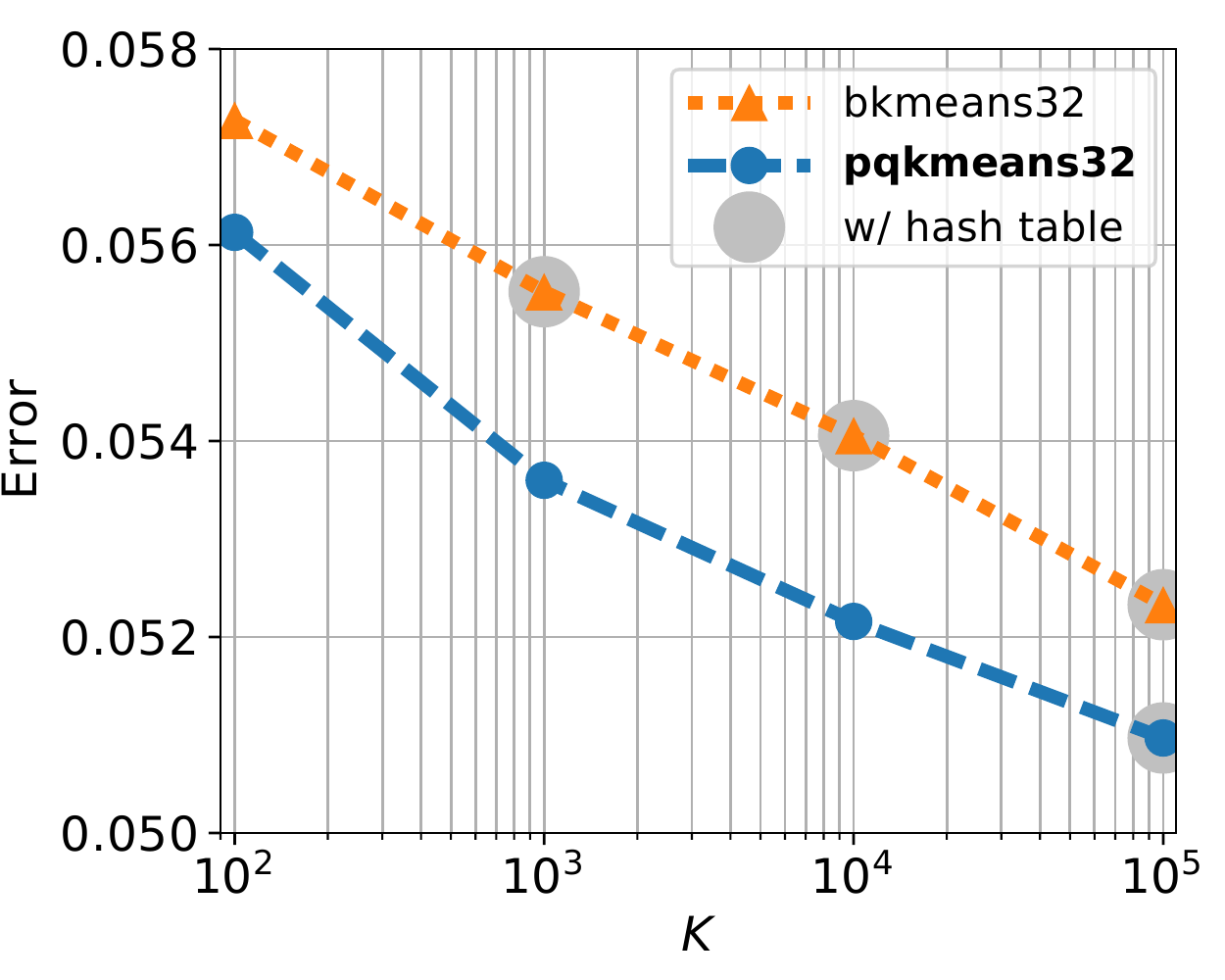}
			\label{fig:large_exp_yfcc100m_k_error}}
		\subfloat[Time according to $K$]{\includegraphics[width=\large_exp_yfcc100m_width]{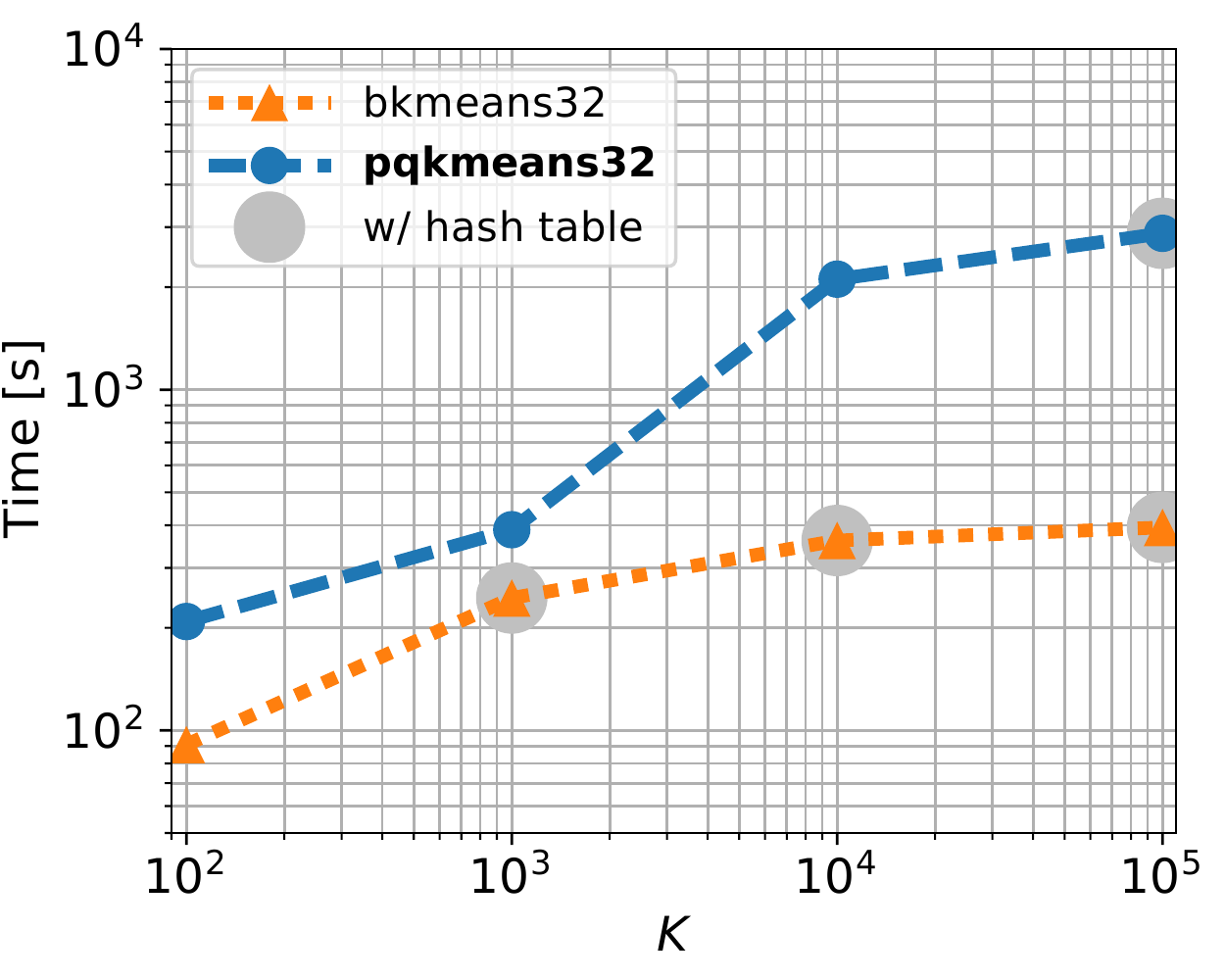}
			\label{fig:large_exp_yfcc100m_k_time}}
	\end{center}
		\vspace{-2mm}
	\caption{Large-scale clustering evaluation for the YFCC100M dataset with five iterations.}
	\label{fig:large_exp_yfcc100m}
		\vspace{-3mm}
\end{figure}

\textbf{YFCC100M:}
\Fref{fig:large_exp_yfcc100m} presents a comparison between PQk-means and Bk-means.
As discussed in \Sref{sec:exp_comp} and \Sref{sec:exp_comp_several},
PQk-means always achieved more accurate clustering and Bk-means was always faster for the same $K$.

\begin{figure}
	\begin{minipage}{1.0\linewidth}
		\begin{center}
			\def\width5703{0.125}
			\subfloat[]{\includegraphics[height=\width5703\linewidth]{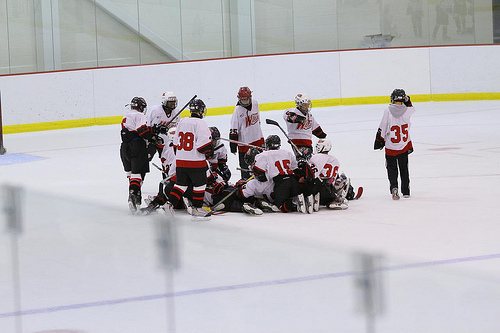}}~
			\subfloat[]{\includegraphics[height=\width5703\linewidth]{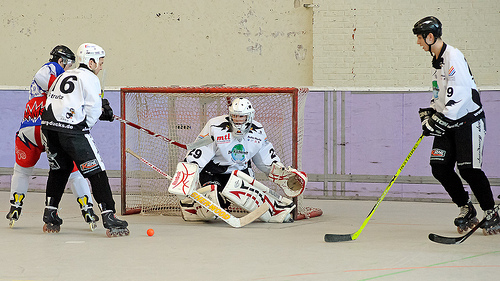}}~
			\subfloat[]{\includegraphics[height=\width5703\linewidth]{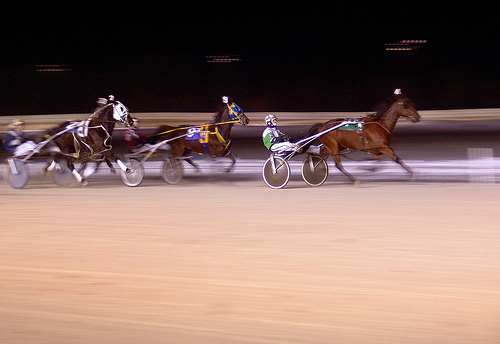}}~
			\subfloat[]{\includegraphics[height=\width5703\linewidth]{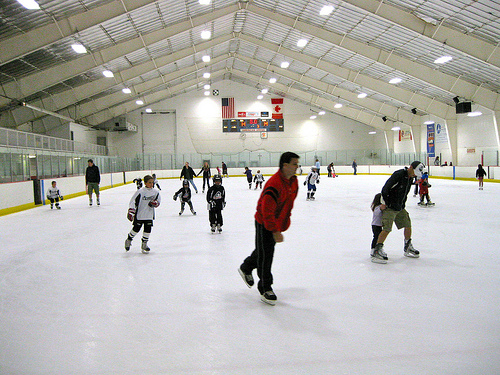}}~
			\subfloat[]{\includegraphics[height=\width5703\linewidth]{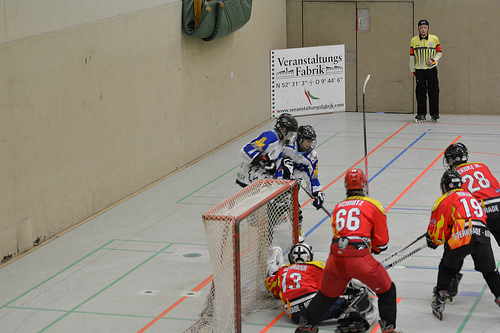}} \\
			\vspace{-3mm}Cluster id: 5703 (a sports game on ice)
		\end{center}
	\end{minipage}
	\vspace{-2mm}
	
	\begin{minipage}{1.0\linewidth}
		\begin{center}		
			\def\width95307{0.14}	
			\subfloat[]{\includegraphics[height=\width95307\linewidth]{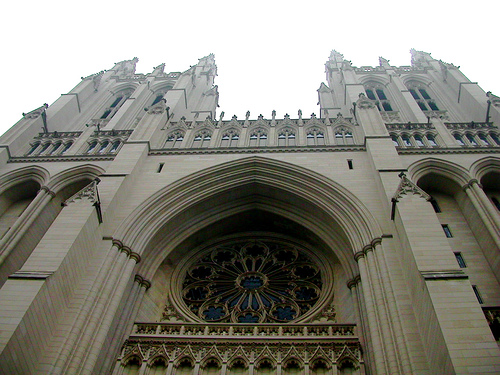}}~
			\subfloat[]{\includegraphics[height=\width95307\linewidth]{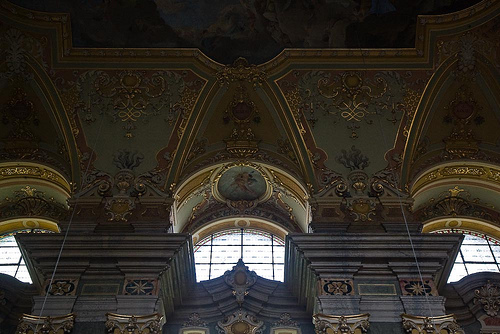}}~
			\subfloat[]{\includegraphics[height=\width95307\linewidth]{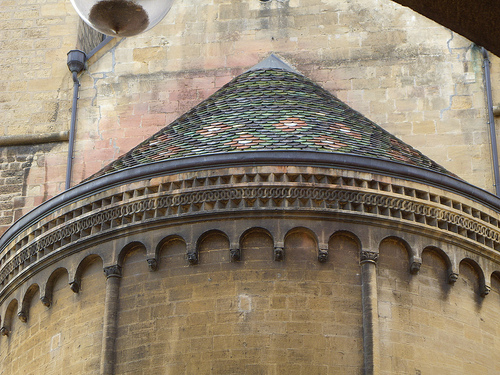}}~
			\subfloat[]{\includegraphics[height=\width95307\linewidth]{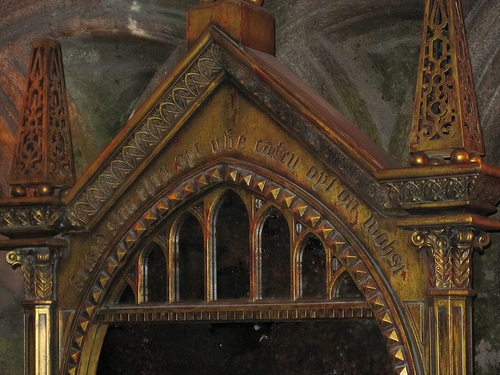}}~
			\subfloat[]{\includegraphics[height=\width95307\linewidth]{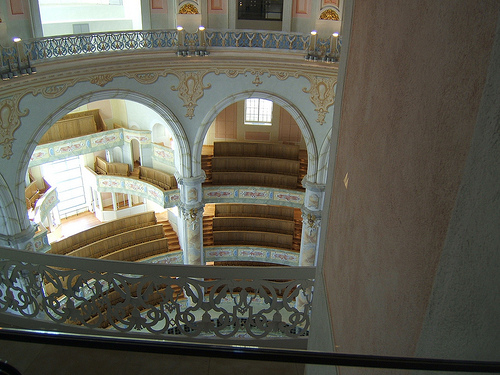}} \\
			\vspace{-3mm}Cluster id: 95307 (a European-style church)	
		\end{center}
	\end{minipage}
	\vspace{-2mm}
	
	
	\begin{minipage}{1.0\linewidth}
		\begin{center}		
			\def\width17713{0.155}	
			\subfloat[]{\includegraphics[height=\width17713\linewidth]{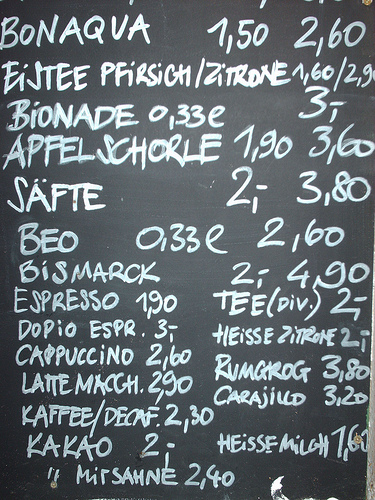}}~
			\subfloat[]{\includegraphics[height=\width17713\linewidth]{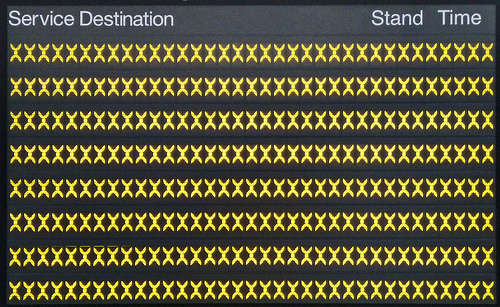}}~
			\subfloat[]{\includegraphics[height=\width17713\linewidth]{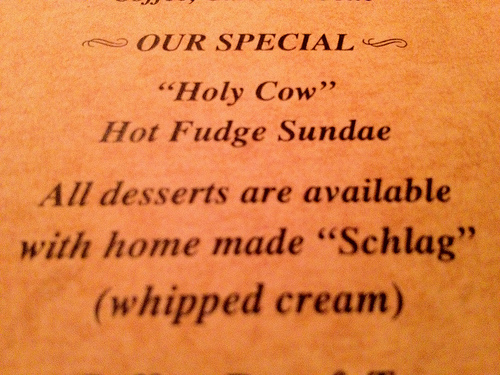}}~
			\subfloat[]{\includegraphics[height=\width17713\linewidth]{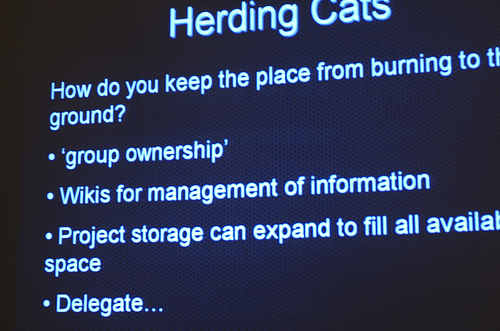}}~
			\subfloat[]{\includegraphics[height=\width17713\linewidth]{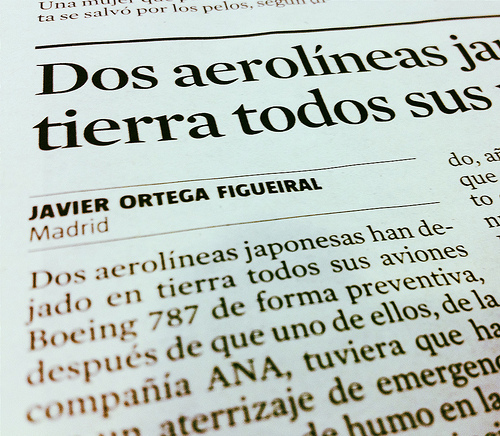}} \\
			\vspace{-3mm}Cluster id: 17713 (some texts)	
		\end{center}
	\end{minipage}
	\vspace{-2mm}

	\begin{minipage}{1.0\linewidth}
		\begin{center}		
			\def\width9566{0.165}	
			\subfloat[]{\includegraphics[height=\width9566\linewidth]{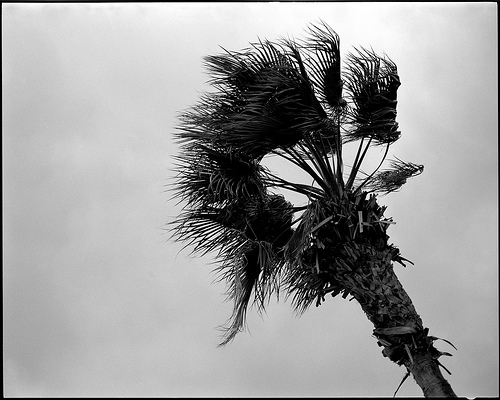}}~
			\subfloat[]{\includegraphics[height=\width9566\linewidth]{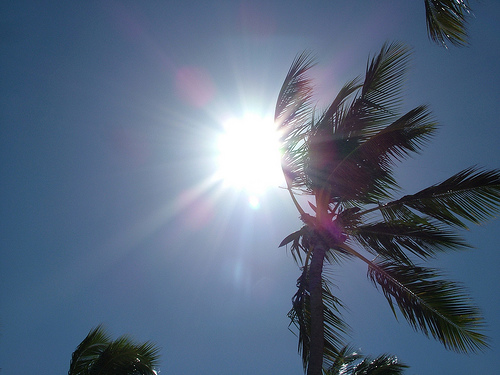}}~
			\subfloat[]{\includegraphics[height=\width9566\linewidth]{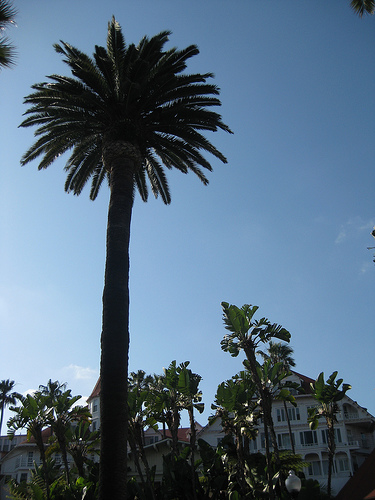}}~
			\subfloat[]{\includegraphics[height=\width9566\linewidth]{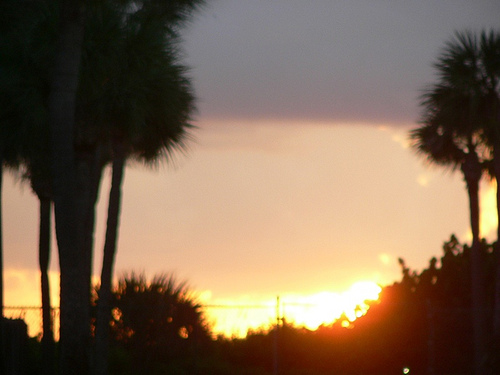}}~
			\subfloat[]{\includegraphics[height=\width9566\linewidth]{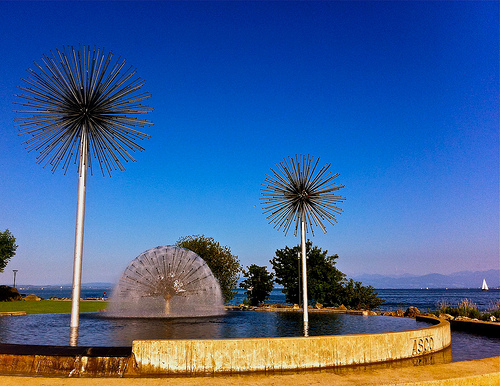}} \\
			\vspace{-3mm}Cluster id: 9566 (a palm tree)			
		\end{center}
	\end{minipage}	
	\vspace{-2mm}
	
	\begin{minipage}{1.0\linewidth}
		\begin{center}		
			\def\width76803{0.15}	
			\subfloat[]{\includegraphics[height=\width76803\linewidth]{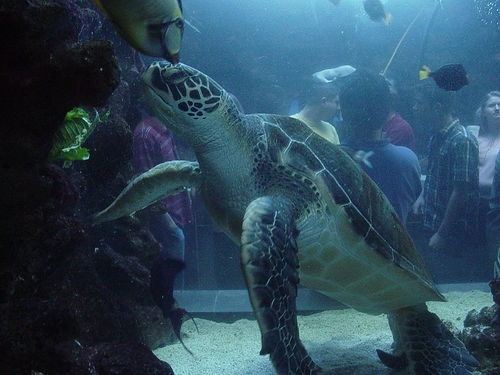}}~
			\subfloat[]{\includegraphics[height=\width76803\linewidth]{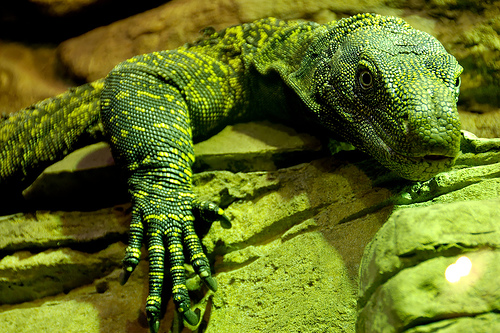}}~
			\subfloat[]{\includegraphics[height=\width76803\linewidth]{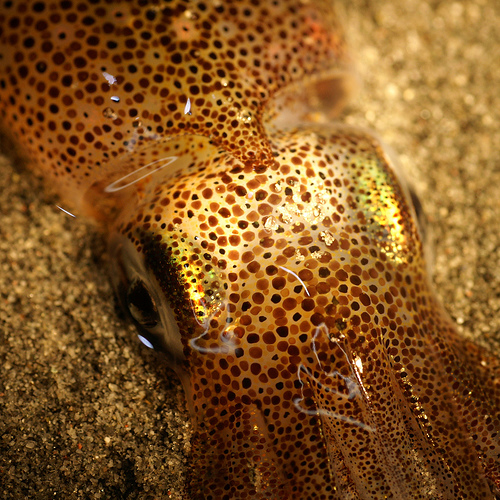}}~
			\subfloat[]{\includegraphics[height=\width76803\linewidth]{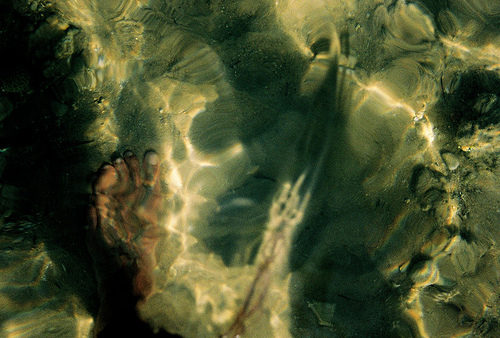}}~
			\subfloat[]{\includegraphics[height=\width76803\linewidth]{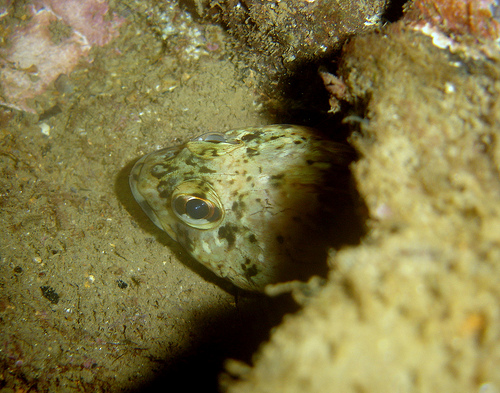}} \\
			\vspace{-3mm}Cluster id: 76803 (a creature of the sea)		
		\end{center}
	\end{minipage}
	\vspace{-2mm}
	
	\caption{Example images from image clustering using PQk-means with $B=32$ and $K=10^5$ for the YFCC100M dataset. Each row shows images belonging to the same cluster.}
	\label{fig:cluster_example}
	\vspace{-3mm}
\end{figure}

The resulting images for the clustering with PQk-means with $B=32$ and $K=10^5$ are presents in \Fref{fig:cluster_example}.
These results show that PQk-means successfully clustered the images.
The images in each cluster show a consistent scenario, as follows:
cluster ID 5703 shows a sports game on ice,
cluster ID 95307 a European-style church,
cluster ID 17713 some texts,
cluster ID 9566 a palm tree, and
cluster ID 76803 a sea creature.
From these results, we conclude that clustering using only 32-bit codes can provide useful results.

\captionsetup[subfigure]{labelformat=empty}

\textbf{SIFT1B and Deep1B:}
\Tref{tbl:large_exp_sift1b_deep1b} presents the runtime evaluation for SIFT1B and Deep1B.
Remarkably, the runtime results for SIFT1B and Deep1B 
exhibit similar behavior, even though the data distribution of SIFT features and GoogLeNet features would be different.
These results indicate that we can predict the runtime performance of PQk-means.
This is important, because estimating the runtime of large-scale clustering is usually difficult.

Although the required memory was less than 32 GB,
PQk-means can handle $10^9$ vectors with $K=10^5$ in around just half a day
(14 hours for SIFT1B and 12 hours for Deep1B).
This implies that PQk-means allows practical large-scale clustering on a single machine.

\begin{table}
	\begin{center}
		\caption{Large-scale clustering evaluation of PQk-means for the SIFT1B and Deep1B datasets with five iterations ($B=32$).}
		\label{tbl:large_exp_sift1b_deep1b}				
		\begin{tabular}{@{}lllllc@{}} \toprule
			Dataset & $N$ & $K$ & Error & Time [s] & w/ table \\ \midrule
			\multirow{4}{*}{SIFT1B} & \multirow{4}{*}{$10^9$} & $10^2$ & 303.4 & $1.88\times10^3$ (31 m) & \\
			& & $10^3$ & 277.4 & $3.95\times10^3$ (66 m) & \\
			& & $10^4$ & 256.0 & $3.68\times10^4$ (10 h) & \\
			& & $10^5$ & 235.1 & $5.14\times10^4$ (14 h) & $\checkmark$ \\ \midrule
			\multirow{4}{*}{Deep1B} & \multirow{4}{*}{$10^9$} & $10^2$ & 0.800 & $1.98\times10^3$ (33 m) & \\
			& & $10^3$ & 0.741 & $4.04\times10^3$ (67 m) & \\
			& & $10^4$ & 0.697 & $3.68\times10^4$ (10 h) & \\
			& & $10^5$ & 0.655 & $4.47\times10^4$ (12 h) & $\checkmark$ \\ \bottomrule
		\end{tabular}
	\end{center}
	\vspace{-5mm}
\end{table}

\subsection{Discussions}
\label{sec:discussion}
\textbf{Comparison with Bk-means:}
The comparative studies illustrated
that both PQk-means and Bk-means are less accurate than the original k-means method.
However, they are both considerably faster, and use significantly less memory.

PQk-means was more accurate than Bk-means in all settings.
Remarkably, PQk-means with 32-bit codes sometimes achieved a better accuracy than Bk-means with 64-bit codes
(Figs.~\ref{fig:time_error_ilsvrc100c}, \ref{fig:time_error_ilsvrc1000c}, and \ref{fig:memory_error}).
In terms of the computational cost, Bk-means was faster than PQk-means,
especially for large $K$.
This difference stems from the fast search using 
hash tables~\cite{tpami_norouzi2014},
which was faster than using the PQTable~\cite{iccv_matsui2015} for PQ codes.
The next step should be to improve this assignment step using an even more efficient data structure. 

An important advantage of PQ codes is that the original vectors can be approximately
reconstructed from the PQ codes.

\textbf{Comparison with Ak-means:}
Compared with PQk-means for the same value of $K$, Ak-means achieved lower errors.
However, it was slower, especially for relatively small values of $K$ (\Fref{fig:k_time}, \Fref{fig:n_time}) or large values of $D$ (\Tref{tbl:exp_comp}).
Because Ak-means stores the original $D$-dimensional vectors,
it requires significantly more memory space than PQk-means.
The advantage of Ak-means is that it does not require an encoding step.
Ak-means would be useful for relatively small-scale problems,
where all of the original vectors can be stored in the memory.


\textbf{Comparison with IQ-means:}
IQ-means is an accelerated version of ranked-retrieval~\cite{wsdm_broder2014}
that skips distance computations when vectors are placed far away from centers.
IQ-means can be the fastest clustering method for large-scale data.
However, IQ-means seems not memory-efficient, and its accuracy was much lower than PQk-means for the YFCC100M dataset.
Please refer to our supplementary material for the discussion on IQ-means.

\section{Conclusions}
In this paper, we introduced the PQk-means clustering method, which is a billion-scale clustering
algorithm for PQ codes.
The proposed method consists of two steps: an assignment step using a PQTable and an update step with a sparse-voting scheme.
PQk-means can cluster even high-dimensional vectors efficiently, because the runtime and memory cost do not depend on the dimensions of the original vectors.
For the same code length, the accuracy of PQk-means was shown to be consistently superior to that of Bk-means, with additional an computational cost.
Experimental results demonstrated that the PQk-means achieved billion-scale clustering within around half a day.

The next step will be to boost PQk-means by using GPUs.
Among the widespread applications of GPU, GPU-based acceleration is becoming a promising method for large-scale clustering~\cite{corr_johnson2017}.
Because PQk-means is simple and easy to parallelize,
its performance can be boosted using GPUs.

\textbf{Acknowledgments:}
This work was supported by JST ACT-I Grant Number JPMJPR16UO, Japan.

\bibliographystyle{ACM-Reference-Format}
\balance
\bibliography{sigproc} 

\end{document}